%% file: main.tex
\documentclass[10pt,twocolumn,letterpaper]{article}

\usepackage[pagenumbers]{cvpr} 


\definecolor{cvprblue}{rgb}{0.21,0.49,0.74}
\usepackage[pagebackref,breaklinks,colorlinks,allcolors=cvprblue]{hyperref}

\usepackage{marvosym}
\usepackage{graphicx}
\usepackage{booktabs}
\usepackage{multirow}
\usepackage{amssymb}
\usepackage{listings}
\usepackage{wrapfig}
\usepackage{ulem}
\usepackage{nicefrac}       
\usepackage{microtype}      
\usepackage{pifont}
\usepackage{wrapfig}
\definecolor{linkcolor}{RGB}{255,0,0}
\definecolor{urlcolor}{RGB}{255,105,180}
\definecolor{citecolor}{RGB}{66,168,235}
\hypersetup{colorlinks=true,linkcolor=linkcolor,urlcolor=urlcolor,citecolor=citecolor}

\def\paperID{5915} 
\def\confName{CVPR}
\def\confYear{2025}

\title{HarmonySet: A Comprehensive Dataset for Understanding\\ Video-Music Semantic Alignment and Temporal Synchronization}

\author{
\textbf{Zitang Zhou}$^{\ast,\text{1}, \text{2}}$, \textbf{Ke Mei}$^{\ast,\text{1}}$, 
\textbf{Yu Lu}$^{\text{1},\text{3},}$\textsuperscript{\Letter},
  \textbf{Tianyi Wang}$^{\text{1}}$, 
  \textbf{Fengyun Rao}$^{\text{1}}$\\
  $^{\text{1}}$ WeChat Vision, Tencent Inc. 
  $^{\text{2}}$ Beijing University of Posts and Telecommunications \\
  $^{\text{3}}$ Zhejiang University
}

\begin{document}
\twocolumn[{
   \renewcommand\twocolumn[1][]{#1}%
   \maketitle
    \vspace{-35pt}
   \begin{center}
    \centering
    \includegraphics[width=1.0\linewidth]{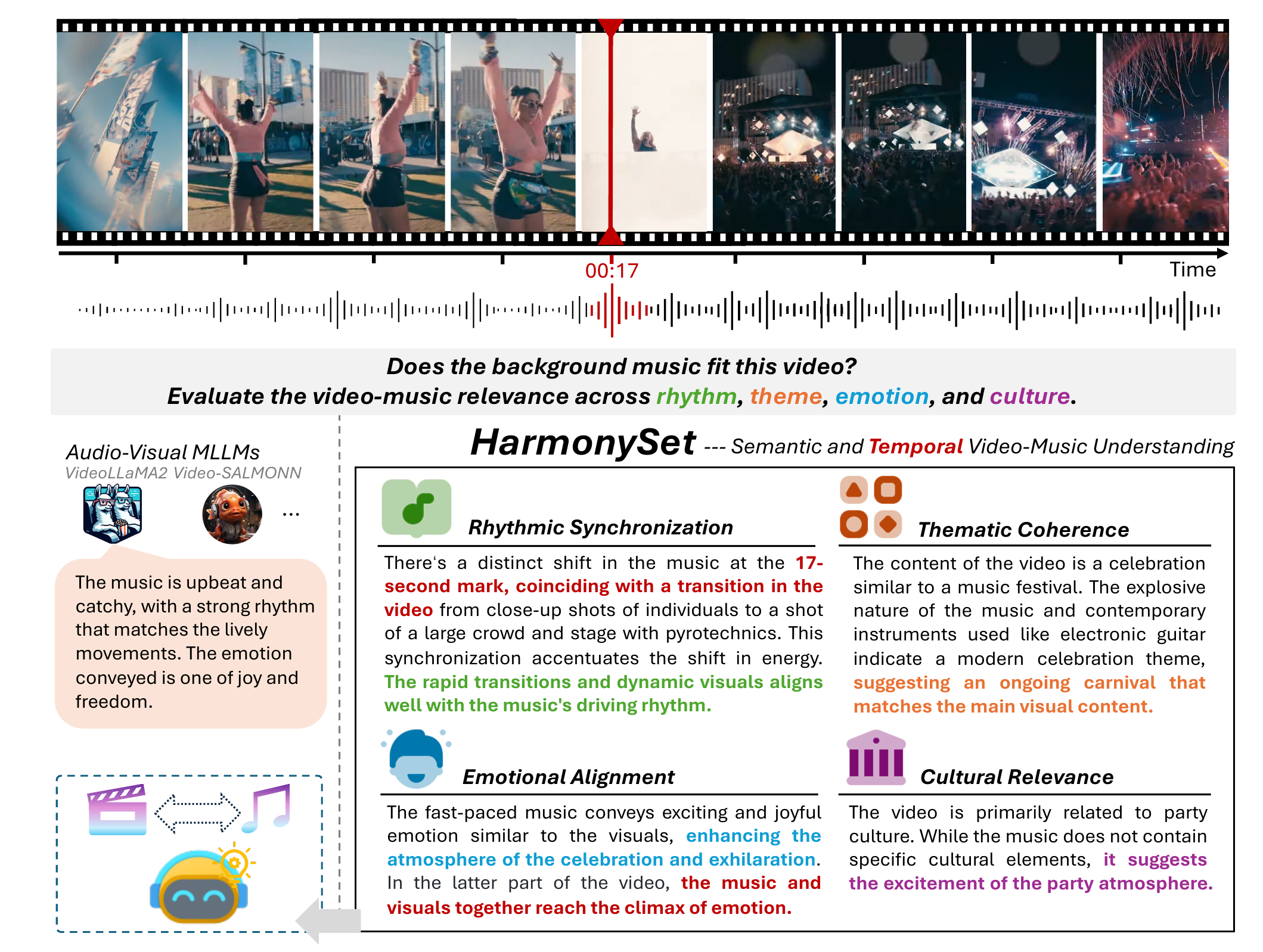}
    \vspace{-10pt}
    \captionof{figure}{We introduce HarmonySet, the first instruction tuning dataset for MLLMs to understand the alignment between video and music. While existing MLLMs typically offer surface-level interpretations of video-music relationships, HarmonySet includes 48,328 video-music pairs, each annotated with rich information on rhythmic synchronization, emotional alignment, thematic coherence, and cultural relevance.}
        \vspace{-5pt}
    \label{fig:motivation}
   \end{center}%
  }]
  %
\renewcommand{\thefootnote}{}
\footnotetext{\textsuperscript{$\ast$} Equal conrtibution \textsuperscript{\Letter} Corresponding author: Yu Lu.}
\renewcommand{\thefootnote}{\arabic{footnote}} 

\input{sec/0_abstract}    
\input{sec/1_intro}

\input{sec/2_related}
\input{sec/3_data}
\input{sec/4_exp}
\input{sec/5_conclusion}

{
    \small
    \bibliographystyle{unsrt}
    \bibliography{main}
}

\clearpage

\input{sec/X_suppl}

\end{document}

%% file: sec/0_abstract.tex
\begin{abstract}
This paper introduces HarmonySet, a comprehensive dataset designed to advance video-music understanding. HarmonySet consists of 48,328 diverse video-music pairs, annotated with detailed information on rhythmic synchronization, emotional alignment, thematic coherence, and cultural relevance. We propose a multi-step human-machine collaborative framework for efficient annotation, combining human insights with machine-generated descriptions to identify key transitions and assess alignment across multiple dimensions. Additionally, we introduce a novel evaluation framework with tasks and metrics to assess the multi-dimensional alignment of video and music, including rhythm, emotion, theme, and cultural context. Our extensive experiments demonstrate that HarmonySet, along with the proposed evaluation framework, significantly improves the ability of multimodal models to capture and analyze the intricate relationships between video and music. Project page: \href{https://harmonyset.github.io/}{https://harmonyset.github.io/}.
\end{abstract}

%% file: sec/1_intro.tex
\section{Introduction}
\label{sec:intro}
\input{table/comparison}

\hspace{\parindent}The rapid growth of online video platforms has led to a rising demand for multimodal content analysis across video, text, and music. This demand is fueled by advancements in large-scale multimodal datasets \cite{bai2023touchstone, li2024mvbench, li2024seed, mangalam2023egoschema, liu2025mmbench, yu2023mm, fu2024video, wu2023q} and models, such as Video Multimodal Large Language Models (MLLMs) \cite{maaz2023video, zhang2023llama, lin2023video, li2024llava, team2024gemini}, which show strong potential for understanding video semantics and performing cross-modal reasoning tasks.

Despite these advancements, a key challenge remains in the domain of video-music understanding, where capturing the complex semantic and temporal relationships between video content and music proves difficult \cite{synchronizationmuvi}. Effective video-music understanding requires the ability to recognize nuanced elements, such as emotional tone, narrative progression, and symbolic imagery—critical aspects that underlie the synchronization between video and music. Current models \cite{cheng2024videollama, sun2024video, chen2023valor, lyu2023macaw}, however, often provide surface-level interpretations of video-music relationships, failing to capture deeper, context-specific insights, such as rhythm synchronization, emotional alignment, and thematic coherence (as illustrated in the left panel of Figure~\ref{fig:motivation}).

A significant limitation in addressing these challenges is the lack of effective datasets that provide comprehensive annotations for video-music understanding. Existing datasets offer paired video and music content \cite{hong2017sspp, li2018creating, tian2024vidmuse, li2022learning, yi2021cross, teng2024mvbind, suris2022s}, but their textual annotations typically consist of basic descriptions \cite{chi2024mmtrail, dong2024musechat, mckee2023language} that fail to capture the detailed semantic alignment and temporal synchronization necessary for effective training of MLLMs. This results in a limited understanding of how music influences the narrative rhythm and emotional tone of video content.

Creating datasets that capture these complex video-music relationships with detailed annotations is a labor-intensive process. Annotators must watch videos while listening to the accompanying music, carefully identifying key transitions to ensure precise temporal alignment. Furthermore, evaluating video-music pairs is inherently subjective \cite{ferraro2021break}, as personal taste and cultural context can significantly influence interpretations, making it difficult to standardize annotations.

To address these challenges, we introduce HarmonySet, a novel dataset designed to facilitate a deeper understanding of video-music alignment. HarmonySet consists of 48,328 diverse video-music pairs, curated from a broad range of genres to ensure comprehensive representation. Each pair is annotated with rich information on key aspects of temporal synchronization and semantic alignment, enabling more robust training of multimodal models. As illustrated in Figure~\ref{fig:motivation}
, HarmonySet provides annotations that go beyond simple descriptions, offering detailed insights into how video and music align both temporally and semantically.

To efficiently generate these annotations, we propose a multi-step human-machine collaborative labeling framework. Initially, human annotators identify key timestamps that mark synchronized transitions between video and music, forming the foundation for deeper analysis. These timestamps serve as anchors for categorizing the video-music alignment into dimensions such as rhythm synchronization, emotional alignment, thematic coherence, and cultural relevance. Annotators then assess each dimension on a scale, ensuring that the annotations capture the full complexity of the video-music relationship. Machine-generated descriptions are subsequently produced by an MLLM \cite{team2024gemini}, which utilizes the identified timestamps and video metadata to provide detailed, context-aware descriptions of the video-music alignment. This combined human-machine approach significantly reduces annotation workload while maintaining high-quality, multi-dimensional insights.

In addition to the dataset, we introduce a novel evaluation framework for benchmarking video-music understanding models. Our framework includes a series of tasks and metrics designed to evaluate critical aspects of video-music alignment, such as temporal synchronization, emotional congruence, and thematic integration. By providing standardized benchmarks, we aim to establish a more rigorous approach to evaluating the performance of models in understanding the complex interplay between video and music.

Comprehensive experiments demonstrate that both HarmonySet and our evaluation framework significantly enhance the ability of multimodal models to capture and analyze the intricate relationships between video and music.

Our key contributions are threefold:
\begin{itemize}
    \item  We introduce HarmonySet, a diverse collection of video-music pairs with rich annotations on rhythmic synchronization, emotional alignment, and thematic coherence, addressing the gap in existing datasets for video-music understanding.
    \item We propose an efficient, multi-step human-machine framework for annotating video-music relationships. This approach combines human insights with machine-generated descriptions to label key transitions and assess alignment across multiple dimensions.
    \item We introduce a new evaluation framework with tasks and metrics for assessing temporal alignment, emotional congruence, and thematic integration, providing a standardized benchmark for video-music understanding tasks.
\end{itemize}

%% file: table/comparison.tex
\begin{table*}[t]
\centering
\caption{\textbf{Overview of Video-Music Datasets.} HarmonySet provides comprehensive video-music content, and stands out among existing video-music datasets by offering both semantic matching and temporal synchronization annotations.}
\small
\resizebox{\linewidth}{!}{
\begin{tabular}{lcccccccccc}
\toprule
\textbf{Dataset}             & \textbf{Year} & \textbf{Music Style} & \textbf{\#Hours} & \textbf{\#Videos}  & \textbf{\#Annotations} & \textbf{Semantic Matching} & \textbf{Temporal Synchronization} \\ \hline
TT-150K \cite{yi2021cross}          & 2021    &  diverse             &    -     &  146,351     &  -    &  \textcolor{red}{\ding{55}}   & \textcolor{red}{\ding{55}}\\
 MovieClips \cite{suris2022s}          &  2022  &   diverse           &   230      &  20,000       &  -    &  \textcolor{red}{\ding{55}}   & \textcolor{red}{\ding{55}}\\
MuseChat \cite{dong2024musechat}      &   2024    &   songs         &     -    &  98,206      &  98,206    &   \textcolor{teal}{\ding{51}}  &\textcolor{red}{\ding{55}}\\
BGM909 \cite{li2024diff}         &    2024   & piano            &    -     &  909      &   9,090   &  \textcolor{red}{\ding{55}}   & \textcolor{red}{\ding{55}}\\
SVM-10K \cite{teng2024mvbind}      &   2024      &  diverse              &    -     &  10,000     &   -  &  \textcolor{red}{\ding{55}}   & \textcolor{red}{\ding{55}} \\
MMTrail \cite{chi2024mmtrail}       &    2024    &   diverse       &     27,100    &  290,000   &  290,000  &   \textcolor{red}{\ding{55}}   & \textcolor{red}{\ding{55}}  \\
    \hline
\textbf{HarmonySet} (Ours) & 2024 &    diverse      &   458.8   &    48,328   & 48,328     & \textcolor{teal}{\ding{51}} & \textcolor{teal}{\ding{51}}\\ \bottomrule
\end{tabular}}
\label{T:related}
\end{table*}

%% file: sec/2_related.tex
\section{Related Work}
\label{sec:related_work}
\subsection{Video-Audio Datasets} Existing datasets used for training MLLMs emphasize general audio features, rather than the specific musical elements that are central to modern video multimodal contents. 
For instance, AudioSet \cite{gemmeke2017audio} and VGGSound \cite{chen2020vggsound} are large-scale datasets primarily designed for audio event recognition. Other datasets like FSD50K \cite{fonseca2021fsd50k} and ESC-50 \cite{piczak2015esc} are also commonly employed in pre-training multimodal models that accept audio inputs. \\
\subsection{Video-Music Datasets} As shown in Table \ref{T:related}, recent benchmarks \cite{hong2017sspp, tian2024vidmuse, li2022learning} incorporate video-music content, exploring video-level visual-music semantic alignment. For instance, TT-150K \cite{yi2021cross} collected 150,000 short videos with music tracks for video-music recommendation. SVM-10K \cite{teng2024mvbind} collected short videos with high likes for filtering high-quality music. MovieClips \cite{suris2022s} comprises 20,000 videos sourced from the MovieClips YouTube channel. However, these datasets merely offer paired music and video data without detailed annotations, limiting their utility in enhancing MLLM capabilities. Some datasets \cite{li2021ai, zhuo2023video, yu2023long, li2024diff, yang2024beatdance, zhu2022quantized, li2018creating} provide annotations for video-music rhythm matching, such as BGM909 \cite{li2024diff} providing short music descriptions, music chords, and beats, but they lack analysis of emotional alignment and semantic transitions. MMTrail \cite{chi2024mmtrail} provides trailer videos and includes descriptions for MLLM instruction tuning, but it does not thoroughly investigate the video-music relationship. Musechat \cite{dong2024musechat} and YT8M-MusicTextClips \cite{mckee2023language} automatically formulated music recommendation dialogues. None of these datasets deliver cohesive and multi-dimensional reasoning on the intricate video-music relationships. The temporal synchronization that enhances the harmony between music and visual narratives remains largely unexplored.

\subsection{Video Datasets and Benchmarks}
Traditional Vision-Language (VL) benchmarks \cite{xu2016msr, xu2017video, goyal2017something, kay2017kinetics, xiao2021next} have primarily focused on specific capabilities such as multimodal retrieval and vision question answering (QA). The advent of multimodal large language models (MLLMs) has spurred the development of benchmarks designed to assess more integrated VL tasks \cite{bai2023touchstone, li2024mvbench, ying2024mmt, zhou2024mlvu, li2024seed, wu2024longvideobench, lu2024exploiting, lu2023show, lu2024zero, xie2024funqa}. For instance, VideoMME \cite{fu2024video}, MM-Vet \cite{yu2023mm}, Q-Bench \cite{wu2023q}, EgoSchema \cite{mangalam2023egoschema}, and MMBench \cite{liu2025mmbench} emphasize comprehensive VL skills. These benchmarks introduce evaluation metrics that go beyond simple model hierarchies, providing a more nuanced assessment of model performance across a range of vision-language tasks.

\subsection{Multimodal Large Language Models}
Video large language models have evolved significantly from captioning tools like BLIP2 \cite{li2023blip} to more advanced systems such as VideoChat \cite{li2023videochat} and Video-LLaVA \cite{lin2023video}, which demonstrate capabilities in dialogue generation and question-answering \cite{yang2024octopus, li2024llava, zhang2023llama, videollama, li2023videochat, maaz2023video}. Increasingly, models are also incorporating audio modalities \cite{chen2023vast, zhu2023languagebind}. Examples include VideoLLaMA2 \cite{cheng2024videollama}, video-SALMONN \cite{sun2024video}, Macaw-LLM \cite{lyu2023macaw}, and VALOR \cite{chen2023valor}, which can analyze both video and audio content and provide open-ended text outputs. These methods leverage powerful language models and can provide a deeper understanding of the relationship between video, audio, and text content, going beyond mere video-audio matching.\\

%% file: sec/3_data.tex
\vspace{-0.5cm}
\section{The HarmonySet Dataset \& Benchmark}

\textbf{HarmonySet} is designed to advance the understanding of video-music relationships by examining how background music aligns with and enhances visual narratives. This dataset emphasizes key aspects of synchronization and semantic alignment, focusing on temporal dynamics, rhythm, theme, emotion, and cultural relevance. In this section, we describe the data collection and annotation process, present dataset statistics, and discuss quality control measures. We demonstrate that \textbf{HarmonySet} is a pioneering resource for studying video-music alignment, offering rich insights into the synchronization between music and visual storytelling.

\subsection{Video Collection}

\begin{figure*}[t]
    \centering
    \includegraphics[width=\textwidth]{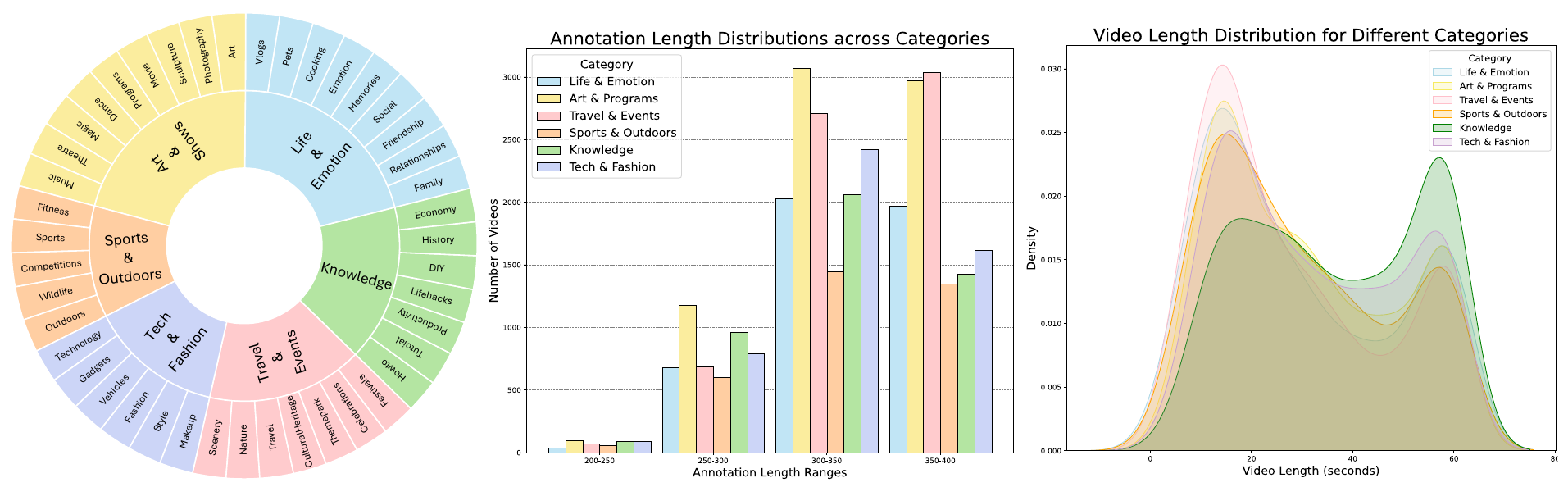} 
    \caption{\textbf{HarmonySet Statistics.} (Left) HarmonySet covers 6 main categories and is divided into 43 subclasses with a full spectrum of content types. (Middle) Distributions of the number of words across categories in HarmonySet annotations. HarmonySet has a balanced annotation length across 6 main categories. (Right) Video duration distributions for different categories. The video durations are concentrated between 10 seconds and 60 seconds, with a rich number of videos in each time segment.}
    \label{fig:Statistics}
\end{figure*}

To ensure a diverse and high-quality collection of video-music pairs, we implemented a hierarchical tagging structure to facilitate the identification of videos that feature well-aligned background music. This structure includes primary categories such as \textit{Life \& Emotions}, \textit{Arts \& Performance}, \textit{Travel \& Events}, \textit{Sports \& Fitness}, \textit{Knowledge}, and \textit{Technology \& Fashion}, each of which represents a broad genre, format, and cultural expression. These categories are further subdivided into 43 specific subcategories (see Figure~\ref{fig:Statistics}, left). In addition, we generated 293 relevant keywords derived from these subcategories to guide our video search process.\\
Using these keywords, we crawled videos from YouTube Shorts, ensuring a variety of music genres and visual content. The dataset exclusively includes videos with user-added background music that complements the visual content. To ensure data consistency, annotators manually reviewed the collected videos to remove those lacking music. To verify the presence of music, we employed the PANNs \cite{kong2020panns} model, which confirmed that 83\% of the videos from our search contained music. Videos without music were excluded to maintain dataset integrity.



\subsection{Annotation Construction}

To make \textbf{HarmonySet} a valuable resource for research on video-music relationships, we implemented a multi-phase annotation process that captures various aspects of the audiovisual content. The annotation process consists of two primary phases: manual annotation by trained annotators and automated refinement using machine-generated annotations.

\subsubsection{Manual Annotation}
Manual annotation includes two main components: synchronization with timestamps and multi-dimensional label assignment.\\
\textbf{Synchronization Annotation}: Annotators identify key moments in the video, such as transitions or shifts in the visual narrative (e.g., scene changes or plot twists). They assess whether the music changes at these points and whether these changes align with the visual transitions, marking the timestamps for temporal synchronization.\\
\textbf{Labeling}: A structured labeling system is used to evaluate the relationship between the video and music across four dimensions: rhythm and synchronization, theme and content, emotion, and cultural relevance. Each label reflects the extent of alignment between the music and video. For example, in the \textit{content alignment} dimension, possible labels include ``strongly related," ``indirectly related," ``unrelated," and ``conflicting." In the \textit{narrative enhancement} dimension, labels include ``enhancing," ``suggesting," ``reversing," ``independent narrative," and ``no supplement." Annotators select the most appropriate label for each dimension, providing a nuanced and multi-faceted understanding of the video-music relationship.\\
Each video is annotated by three independent annotators to ensure objectivity and reliability. The final annotations are derived from the consensus among these annotators, minimizing individual biases and enhancing the robustness of the dataset.

\subsubsection{Quality Control}
A rigorous quality review process was implemented to ensure accurate and reliable annotations. A dedicated reviewer cross-checked each annotation for key timestamp accuracy consistency and factual grounding of the labels. This process mitigates potential biases and ensures data quality.

\subsubsection{Automated Annotation Curation}
Following manual annotation, we employed \textit{Gemini 1.5 Pro} \cite{team2024gemini} to generate enhanced annotations. The inputs for this process included the video and audio content, the manually verified annotations (used as ground truth), and the video metadata (e.g., titles, and descriptions). The system was tasked with generating detailed descriptions of the video-music relationship, focusing on the four key dimensions: rhythm and synchronization, theme and content, emotion, and cultural relevance. The final output provides temporally aligned, multi-dimensional annotations that offer deeper insights into the alignment between video and music. Specific generation prompts and further details on the automated annotation process can be found in the appendix.

\subsection{Instruction Tuning Dataset Statistics}
To further enhance the utility of the HarmonySet dataset for training multimodal models, we created an instruction-tuning dataset. This dataset includes structured annotations that provide detailed explanations of the video-music relationships, enabling the fine-tuning of models for MLLMs to better understand video-music content. \\
The HarmonySet instruction tuning dataset consists of \textbf{44,470} video-music pairs, each with an annotation that provides a structured explanation of the video-music connections. Figure \ref{fig:Statistics} (middle \& right) illustrates the distribution of annotation length and video durations within the HarmonySet dataset. The videos are between 2.96 and 63.38 seconds in length, with an average duration of 31.5 seconds, contributing to a total of \textbf{458.8} hours of video and music content. The average number of words in HarmonySet annotations is \textbf{352.65} words for each video-music pair.

\begin{figure}[t]
    \centering
    \includegraphics[width=\linewidth]{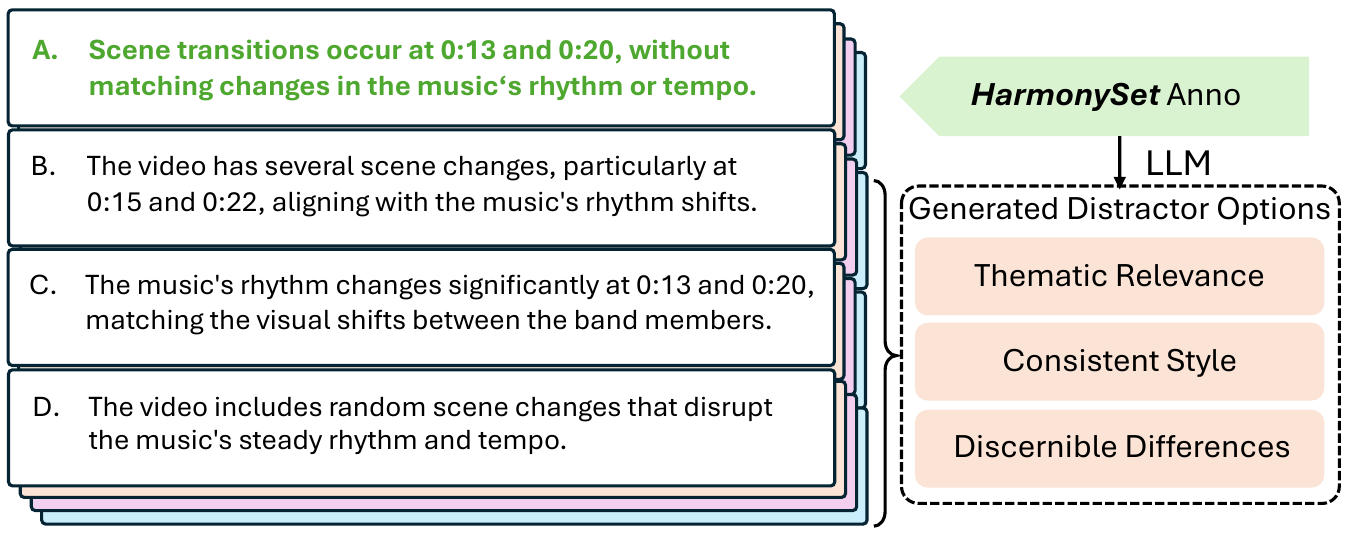} 
    \vspace{-0.5cm}
    \caption{An example of HarmonySet-MC curation. We used LLM to convert open-ended annotations into multiple-choice options, with HarmonySet annotations serving as the correct options. Wrong options are constructed to be challenging yet distinguishable from the correct option.}
    \label{fig:mc}
    \vspace{-0.5cm}
\end{figure}

\subsection{HarmonySet Benchmark}
\textbf{HarmonySet-OE} HarmonySet-OE is comprised of 3,858 video-music pairs along with their accompanying annotations. MLLMs are required to address the video-music alignment relationships, including both temporal synchronization and semantic matching. The expected responses are designed to be open-ended to cover the diverse angles of video-music relationships. Traditional language metrics like BLEU-4 \cite{papineni2002bleu} and ROUGE-L \cite{lin2004rouge} are only sensitive to lexical variations and cannot identify changes in sentence semantics. Recent study \cite{zheng2023judging} has proved LLM \cite{openai2024gpt4technicalreport} to be a reliable evaluation tool for open-ended responses. Therefore, MLLM scores are obtained by comparing the MLLM outputs with the ground truth provided by HarmonySet-OE using LLM. The specific prompt of LLM for evaluation can be found in Appendix.\\
\textbf{HarmonySet-MC} We further developed HarmonySet-MC, a multiple-choice extension of HarmonySet-OE, to facilitate a more structured and objective evaluation process. Specifically, we instructed \textit{GPT-4o} to use the annotated answer as the correct option and create three wrong options. These distracting options were carefully crafted to meet the following criteria: 1) Maintain thematic relevance to the correct answer, avoiding overly obvious discrepancies, 2) Resemble the correct answer in length and sentence structure, avoiding superficial distinctions, and 3) Present discernible semantic differences compared to the correct answer, see Figure \ref{fig:mc} for an example. HarmonySet-MC includes the same 3,858 video-music pairs with HarmonySet-OE, each with four multiple-choice questions corresponding to four distinct aspects of rhythm, emotion, theme, and cultural context. HarmonySet-MC offers a convenient evaluation tool, allowing direct assessment of model performance using multiple-choice accuracy.

\input{table/experiment3}
\subsection{Dataset Assessment}
\begin{figure}[t]
    \centering
    \includegraphics[width=\linewidth]{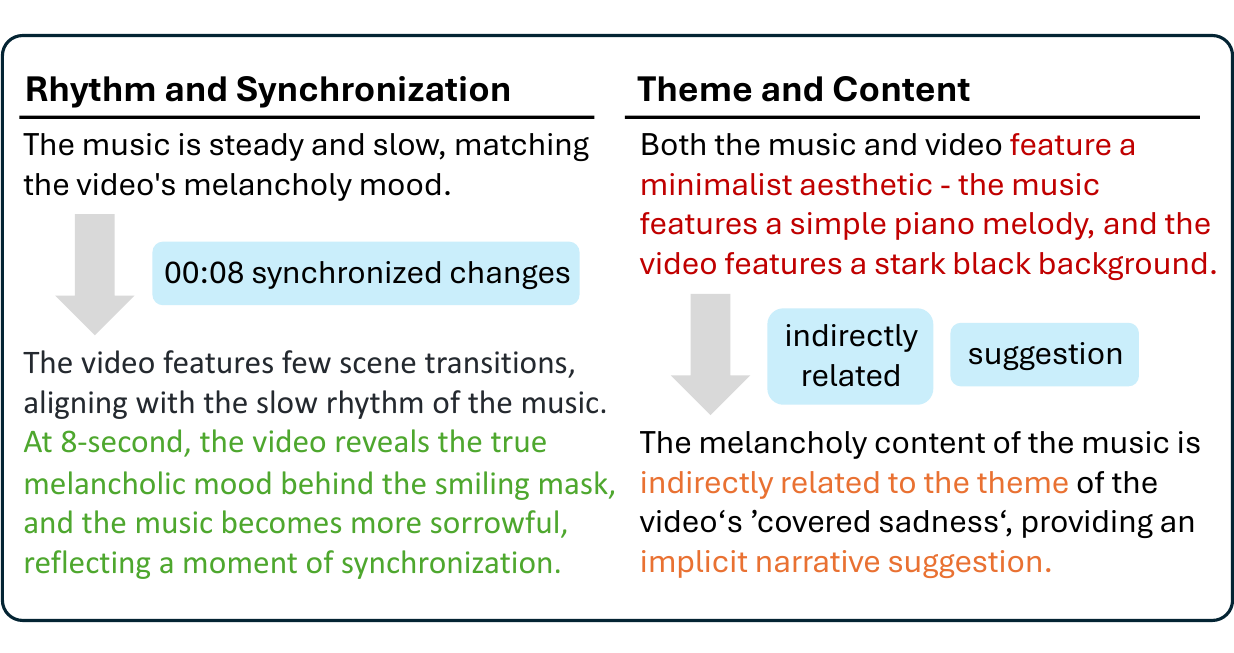} 
    \vspace{-0.5cm}
    \caption{An example of annotation before and after the introduction of manual labels. The red text highlights an unreasonable explanation that arises in the absence of human guidance.}
    \label{fig:anno_exam}
    \vspace{-0.5cm}
\end{figure}

\begin{table}[t]
\centering
\caption{Consensus evaluation result. We randomly sampled 10\% of the data and had human reviewers select ``Low", ``Medium", or ``High" as their level of agreement with the annotations. HarmonySet gets 92\% high consensus, showing a high level of agreement.}
\tiny
\renewcommand\baselinestretch{0.5}
\vspace{-0.2cm}
\resizebox{0.8\linewidth}{!}{
\begin{tabular}{lccc}
\toprule
         & Low                              & Medium                            & High                           \\ \midrule
Consensus   &   3\%  &  5\% &  92\%   \\
 \bottomrule
 \label{T:consensus}
\end{tabular}}
\vspace{-0.5cm}
\end{table}

\noindent Due to the potential influence of individual biases on the annotations, we conducted a consensus evaluation after completing all annotations to ensure agreement and objectivity. The results, detailed in Table \ref{T:consensus}, show that HarmonySet gets 92\% high consensus. \\
\noindent Figure \ref{fig:anno_exam} illustrates the comparison of a video's two aspects of annotation before and after injecting manual labels. Manual annotations provide more accurate semantic alignment understanding and incorporate specific temporal synchronization information. Without human knowledge, models tend to generate spurious video-music connections or meaningless responses. Experiment in Table \ref{T:exp_comparison} also confirms the significant positive impact of human annotation on improving video-music understanding.

\subsection{Dataset Property}
\textbf{HarmonySet emphasis on temporal synchronization.} One crucial factor of video-music connection lies in temporal synchronization. For example, the music becomes more intense just as an athlete makes a final sprint. Such synchronized changes contribute significantly to an immersive visual-music experience. Static images paired with music can only achieve content or mood matching, lacking the ability to express such dynamic transitions. HarmonySet not only includes detailed annotations on the overall pace suitability and beat matching but also provides timestamped explanations of video-music transitions. 58\% of the data in HarmonySet contains key timestamp annotations, providing valuable support for understanding temporal relationships between video and music.\\
\textbf{HarmonySet provides deep semantic alignment understanding.} The semantic resonance between video and music often manifests as a subtle connection that is difficult to articulate. HarmonySet categorizes this semantic alignment into four dimensions, providing a comprehensive framework for understanding these complex relationships.

%% file: table/experiment3.tex
\begin{table*}[t]
\caption{\textbf{Main Results on HarmonySet-OE.} We tested Gemini-1.5 Pro and open source MLLMs including VideoLLaMA2 and video-SALMONN. The bottom part presents results of VideoLLaMA2 finetuned on our instruction tuning dataset. While Gemini-1.5 Pro leads among untrained models, VideoLLaMA2 finetuned with HarmonySet demonstrates significant improvement and a strong understanding of video-music alignment. Results on synchronization can be found in R \& S (Rhythm \& Synchronization), and semantic matching results consist of scores in T (Theme), E (Emotion), and C (Culture).}
\label{T:experiment}
\vspace{-10pt}
{

\normalsize
\tiny
\setlength\tabcolsep{5pt} 
\resizebox{1.0\linewidth}{!}{
\begin{tabular}{lccccccccc}
\toprule

Models   & \begin{tabular}[c]{@{}c@{}}LLM \end{tabular}  & Metrics & \begin{tabular}[c]{@{}c@{}}Life \&\\Emotion \end{tabular}  & \begin{tabular}[c]{@{}c@{}}Art \&\\Performance \end{tabular}  & \begin{tabular}[c]{@{}c@{}}Travel \&\\Events \end{tabular}  &  \begin{tabular}[c]{@{}c@{}}Sports \&\\Outdoors \end{tabular}  & Knowledge &\begin{tabular}[c]{@{}c@{}}Tech \&\\Fashion \end{tabular}  & Overall \\
\midrule
\multicolumn{10}{l}{\textbf{- Close-source MLLM}} \\\midrule
\multirow{4}{*}{Gemini-1.5 Pro \cite{hurst2024gpt}} & \multirow{4}{*}{\textbf{-}} & R \& S   & 5.30 & 6.05 & 5.69 & 4.94 & 4.91 & 4.98 & 5.43 \\
                   &   & T   & 5.25 & 5.76 & 5.75 & 4.41 & 4.46 & 4.49 & \textbf{5.18} \\
                    &   & E & 5.28 & 5.75 & 5.60 & 4.59 & 4.45 & 4.43 & 5.15 \\
                    &   & C & 4.64 & 4.91 & 4.77 & 3.85 & 4.27 & 4.03 & 4.51 \\

\toprule
\multicolumn{10}{l}{\textbf{- Open-source MLLMs}} \\\midrule
\multirow{4}{*}{VideoLLaMA2 \cite{cheng2024videollama}}& \multirow{4}{*}{Qwen2-7B} & R \& S  & 3.89 & 4.80 & 4.56 & 4.01 & 3.39 & 3.54 & 4.15 \\
                    &    & T  & 4.09 & 4.83 & 4.93 & 3.89 & 3.44 & 3.71 & 4.29 \\
                    &    & E & 4.36 & 5.01 & 5.02 & 4.08 & 3.44 & 3.49 & 4.38 \\
                    &    & C & 2.95 & 3.46 & 3.69 & 2.56 & 2.32 & 2.52 & 3.05 \\
\\
\multirow{4}{*}{video-SALMONN \cite{sun2024video}}& \multirow{4}{*}{Vicuna-13B-v1.5} & R \& S   & 2.43 & 3.53 & 2.98 & 2.68 & 2.32 & 2.51 & 2.83 \\
                    &   & T   & 3.24 & 4.18 & 3.97 & 3.23 & 2.96 & 3.00 & 3.55 \\
                    &   & E & 3.11 & 4.12 & 3.84 & 3.13 & 2.56 & 2.70 & 3.38 \\
                    &   & C & 1.85 & 2.51 & 2.51 & 1.77 & 1.68 & 1.84 & 2.12 \\
\toprule
\multicolumn{10}{l}{\textbf{- With HarmonySet}} \\\midrule
\multirow{4}{*}{VideoLLaMA2 (HarmonySet)}& \multirow{4}{*}{Qwen2-7B} & R \& S  & 5.43 & 6.35 & 6.03 & 4.94 & 5.33 & 4.83 & \textbf{5.55} \\
                    &   & T   & 5.12 & 5.21 & 5.03 & 4.84 & 5.21 & 4.85 & 5.06 \\
                    &   & E & 5.25 & 6.41 & 5.84 & 4.00 & 4.88 & 4.47 & \textbf{5.26} \\
                    &   & C & 4.87 & 4.98 & 4.72 & 3.31 & 5.23 & 4.09 & \textbf{4.62} \\
\bottomrule

\end{tabular}}\par
}
\end{table*}

%% file: sec/4_exp.tex
\input{table/exp_comprasion}
\section{Experiments}
\subsection{Baselines}
We conduct the evaluation on Gemini 1.5 Pro \cite{team2024gemini} and state-of-the-art open-source video-audio MLLMs, including VideoLLaMA2 \cite{cheng2024videollama} and video-SALMONN \cite{sun2024video}. For a fair comparison, we adopt the zero-shot setting to infer HarmonySet-OE questions with all MLLMs based on the same prompt. In the experiments presented in Table \ref{T:experiment}, we used a consistent 16 frames for the video input of open-source models for both inference and fine-tuning. A special case is Gemini 1.5 Pro, which supports relatively long multimodal contexts, and videos are sampled at 1 frame per second for the input. In the Appendix, we provide detailed information regarding the architecture and the parameter size for all open-source MLLMs evaluated in this paper, as well as additional results for more MLLMs under various settings. 
\begin{table}[t]
\centering
\caption{Human and model performance on HarmonySet-MC. While VideoLLaMA2 tuned on HarmonySet surpasses Gemini-1.5 Pro in certain aspects, it still falls short of human performance, highlighting both the challenging nature of our task and the limitations of current models.}
\tiny
\resizebox{\linewidth}{!}{
\begin{tabular}{lcccc}
\toprule
         & R \& S (Acc.)                              & T (Acc.)                             & E (Acc.)    & C (Acc.)                          \\ \midrule
Gemini-1.5 Pro   &   41.84\%  &  45.45\% &  44.43\%& 50.40\%   \\
Video-LLaMA2   &   21.76\%  & 48.95\% & 52.76\%  & 24.29\%   \\
Video-LLaMA2 (HarmonySet)   &   10.63\%  &  54.16\% &  47.32\%& 36.66\%   \\
Human & \textbf{85.26\%}  & \textbf{88.19\%} & \textbf{84.49\%} & \textbf{93.81\%} \\
\bottomrule
\end{tabular}}
\label{T:human_performance}
\end{table}
\begin{table}[t]
\centering
\caption{Results of VideoLLaMA2 trained on HarmonySet with 16, 32, and 64 frames. Using 64 frames yields the lowest scores, indicating potential redundancy or even negative effects from excessive visual input within short (\textless1 minute) videos.}
\tiny
\renewcommand\baselinestretch{0.8}
\resizebox{0.8\linewidth}{!}{
\begin{tabular}{lcccc}
\toprule
         & R \& S                              & T                             & E    & C                          \\ \midrule

16 Frames   &   5.55  &  5.06 &  5.26 & 4.62   \\

32 Frames & \textbf{5.59}  & \textbf{5.08} & \textbf{5.29} & \textbf{4.65} \\

64 Frames & 5.49  & 4.94 & 5.21 & 4.53 \\
 \bottomrule
\end{tabular}}
\vspace{-0.5cm}
\label{T:frame_ablation}
\end{table}
\subsection{Main Results}
Table \ref{T:experiment} shows the main results on HarmonySet-OE. Gemini-1.5 Pro generally outperforms untrained open-source MLLMs across all categories and metrics, significantly exceeding the second-best model in Rhythm \& Synchronization (by 1.28) and Culture (by 1.46). This might be due to its capacity for long context inputs and timestamped outputs, allowing for better alignment with human annotations that often consider temporal relationships. The performance in culturally nuanced pairings might stem from Gemini-1.5 Pro's extensive training data.\\
Untrained VideoLLaMA2 and video-SALMONN both underperform Gemini-1.5 Pro. While VideoLLaMA2 shows moderate rhythmic synchronization (4.15), its semantic matching capabilities are weaker, particularly in cultural understanding (3.05). This suggests a deficiency in comprehending nuanced cultural differences and contextual information. Video-SALMONN consistently scores lowest across all metrics and categories, struggling with understanding temporal synchronization, emotional congruence, thematic integration, and cultural relevance. The open-source models' weaker performance likely stems from having less training data (in both quantity and quality, especially regarding cultural nuances) and limited input capacity, hindering analysis of complex relationships and rhythmic synchronization requiring longer temporal contexts.\\
Training VideoLLaMA2 on HarmonySet yields substantial improvements, boosting its Rhythm \& Synchronization score by 1.40 and Culture score by 1.57. This surpasses previous state-of-the-art results in most domains, demonstrating HarmonySet's effectiveness in addressing limitations of models trained on lower-quality data and fostering deeper multimodal understanding.\\
\subsection{HarmonySet-MC Results}
We evaluated different models on HarmonySet-MC and invited individuals unfamiliar with the dataset to provide human performance. From Table \ref{T:human_performance}, it is evident that human performance significantly outperforms the best model results across all evaluation perspectives. This highlights the challenging nature of our dataset, which effectively measures the gap between model performance and human-level understanding. These results indicate that current models still struggle to effectively understand the complex interplay between video and music, underscoring the need for further advancements in model training and dataset annotation to bridge this performance gap. The complete experimental results on HarmonySet-MC can be found in the Appendix. 
\begin{figure*}[t]
    \centering
    \includegraphics[width=\textwidth]{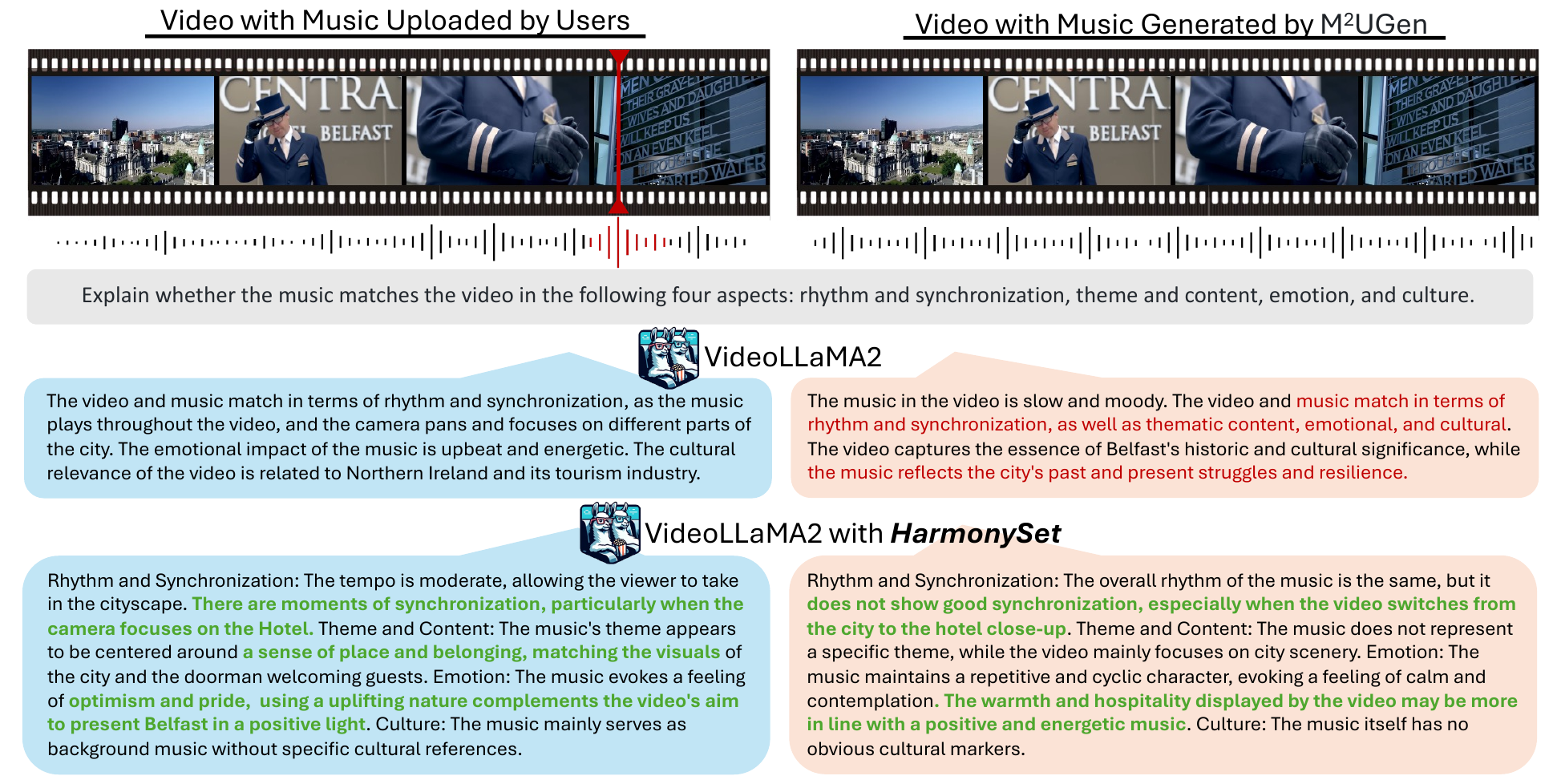} 
    \caption{VideoLLaMA2's response before and after training with our instruction tuning dataset. The left video features human-composed soundtracks, while the right video is with AI-generated soundtracks. Without HarmonySet, the model often provides the wrong justification for the generated music for its harmony with the visual content (highlighted in red text). The trained model offers more insightful analysis and can effectively assess both human-composed and AI-generated music. Our dataset facilitates a deeper understanding of both synchronization and semantic alignment.}
    \label{fig:generation}
    \vspace{-0.3cm}
\end{figure*}
\subsection{Ablations}
\textbf{Ablation on frame number.} We designed experiments to investigate the impact of the number of frames used during training on the model's performance, as shown in Table \ref{T:frame_ablation}. Increasing the frames from 16 to 32 improves the model's performance across all aspects. However, using 64 frames yields the worst performance, even lower than with 16 frames. This might be attributed to the focus of our videos on short-form content under one minute, where 64 frames could introduce increased computational complexity, potential overfitting, or information redundancy.\\
\textbf{Can fully automated data provide sufficient capability?} HarmonySet provides valuable video-music annotations that enhance the multimodal understanding of MLLMs. To evaluate the dataset's effectiveness, we compared two types of training data: annotations fully generated by Gemini 1.5 Pro and those derived from our HarmonySet. Both training processes utilized 10,000 samples. Results show in Table \ref{T:exp_comparison} that training on the fully automated data yields minimal performance gains. Models trained with HarmonySet annotations consistently surpass those trained with auto-generated data, especially in capturing synchronization and semantic alignment between video and music. These results highlight the importance of a human-machine collaborative framework in enhancing multimodal models' video-music understanding. \\

\subsection{Effectiveness on Assessing AI-Generated Music}
Generating music for videos \cite{liu2023m, su2021does, guo2024gotta, mei2024foleygen, liu2024tell} is a highly challenging task that requires harmony between music and visual narratives in terms of synchronization and semantic alignment. Figure \ref{fig:generation} shows VideoLLaMA2's improved ability to differentiate between human-composed and AI-generated soundtracks \cite{liu2023m} after training with HarmonySet. The vanilla VideoLLaMA2 struggled to justify the generated music's harmony with the video, likely due to a lack of understanding of what constitutes harmonious alignment. With HarmonySet, the model provides more detailed temporal and semantic analyses, enabling more objective evaluation. These results offer valuable insights for future video soundtrack creation.


%% file: table/exp_comprasion.tex
\begin{table*}[t]
\caption{Comparison between performance of VideoLLaMA2 trained on fully automated data and HarmonySet data. Models trained with our instruction tuning data demonstrates a clear advantage, validating the value of human expertise in providing rich information on synchronization and semantic alignment and the effectiveness of our human-machine collaborative framework.}
\label{T:exp_comparison}
\vspace{-10pt}
{
\renewcommand\baselinestretch{0.9}

\normalsize
\tiny
\setlength\tabcolsep{5pt} 
\resizebox{1.0\linewidth}{!}{
\begin{tabular}{lcccccccc}
\toprule

Models    & Metrics & \begin{tabular}[c]{@{}c@{}}Life \&\\Emotion \end{tabular}  & \begin{tabular}[c]{@{}c@{}}Art \&\\Performance \end{tabular}  & \begin{tabular}[c]{@{}c@{}}Travel \&\\Events \end{tabular}  &  \begin{tabular}[c]{@{}c@{}}Sports \&\\Outdoors \end{tabular}  & Knowledge &\begin{tabular}[c]{@{}c@{}}Tech \&\\Fashion \end{tabular}  & Overall \\
\midrule

\multirow{4}{*}{VideoLLaMA2 (F.A., 10k)} &  R \& S   & 4.56 & 5.05 & 4.97 & 4.28 & 4.20 & 4.17 & 4.59 \\
                      & T   & 4.20 & 5.01 & 4.85 & 3.49 & 3.36 & 3.41 & 4.16 \\
                       & E & 4.29 & 4.76 & 4.67 & 4.03 & 3.76 & 3.79 & 4.28 \\
                       & C & 3.53 & 3.98 & 3.67 & 2.93 & 3.13 & 3.02 & 3.44 \\

\midrule

\multirow{4}{*}{VideoLLaMA2 (HarmonySet, 10k)}&  R \& S  & 4.69 & 5.58 & 5.30 & 4.49 & 4.36 & 4.25 & \textbf{4.86} \\
                        & T  & 4.66 & 5.02 & 4.98 & 4.40 & 4.39 & 4.29 & \textbf{4.70} \\
                        & E& 4.64 & 5.43 & 5.26 & 4.06 & 3.85 & 3.78 & \textbf{4.66} \\
                       & C & 3.99 & 4.30 & 4.25 & 2.97 & 3.79 & 3.34 & \textbf{3.89} \\
\bottomrule

\end{tabular}}\par
}
\end{table*}

%% file: sec/5_conclusion.tex
\section{Conclusion}
We introduce HarmonySet, the first dataset focused on facilitating the ability of MLLMs in comprehensive video-music understanding. HarmonySet comprises a diverse domain of videos with high-quality music, each annotated with structured explanations detailing the semantic matching and temporal synchronization between video and music. Our extensive evaluation of state-of-the-art MLLMs, encompassing commercial and open-source models, reveals the limitations of MLLMs' reasoning about the interplay between visual and musical elements. This highlights the challenge of achieving in-depth video-music understanding. There are many exciting directions to build upon this work, including developing novel MLLM architectures specifically tailored for video-music analysis and investigating the potential for cross-modal knowledge transfer between video and music content. We hope HarmonySet will inspire future research and development in improving the capabilities of MLLMs.

%% file: sec/X_suppl.tex
\clearpage
\setcounter{page}{1}
\maketitlesupplementary
\appendix

\section{More details of HarmonySet}
\subsection{Dataset Construction pipeline}
\begin{figure*}[t]
    \centering
    \includegraphics[width=\textwidth]{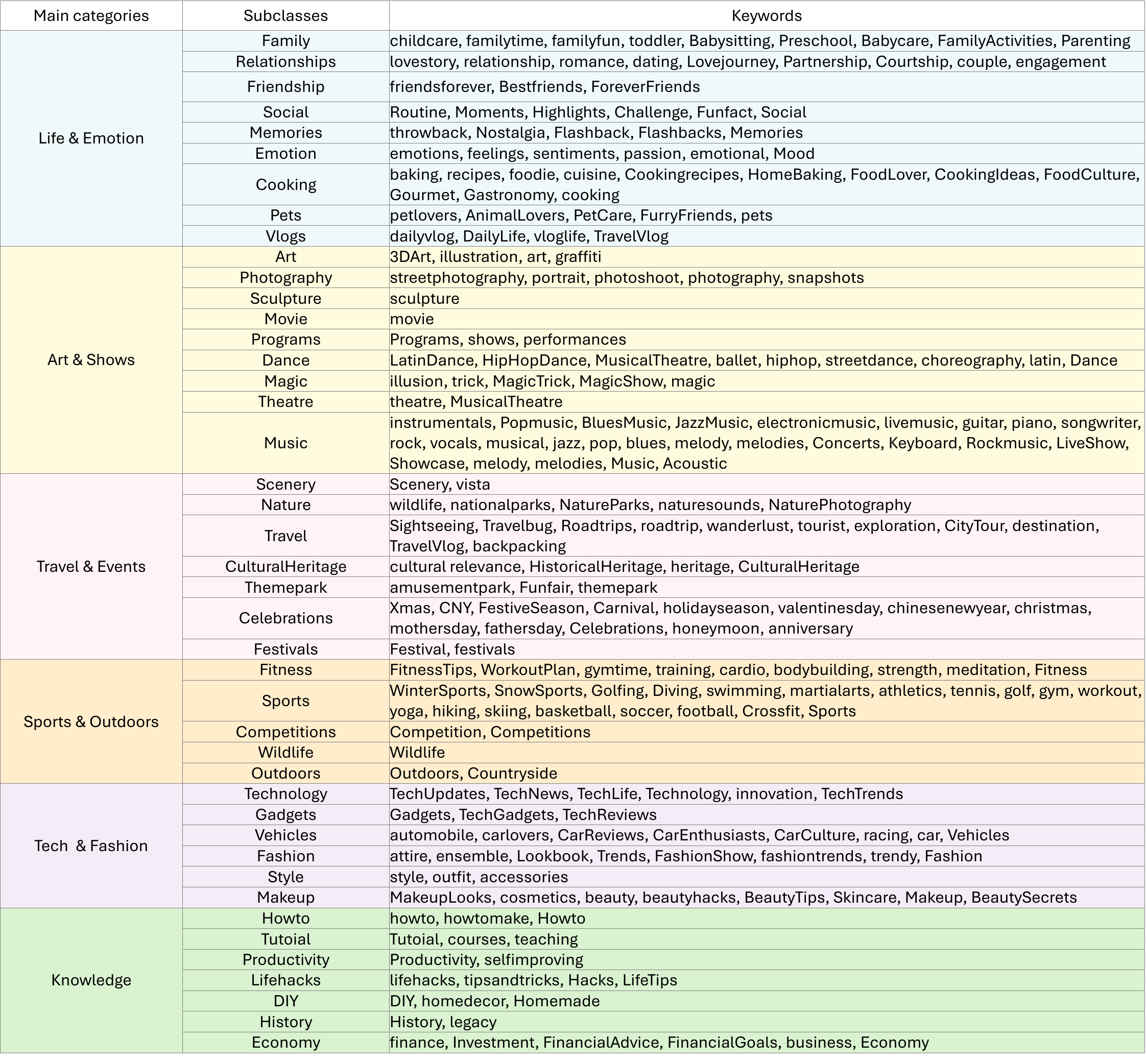} 
    \caption{We employed a hierarchical keyword taxonomy for video collection, comprising six primary categories and 43 subcategories, yielding a total of 293 unique keywords. This taxonomy was meticulously designed to target videos with high-quality music. Keywords unrelated to music, such as \textit{news broadcast} and \textit{read}, were excluded to enhance the precision of the search and prioritize music-centric content. The resulting keyword set was then utilized to crawl and curate a collection of videos.}
    \label{fig:keywords}
\end{figure*}

\begin{figure*}[t]
    \centering
    \includegraphics[width=\textwidth]{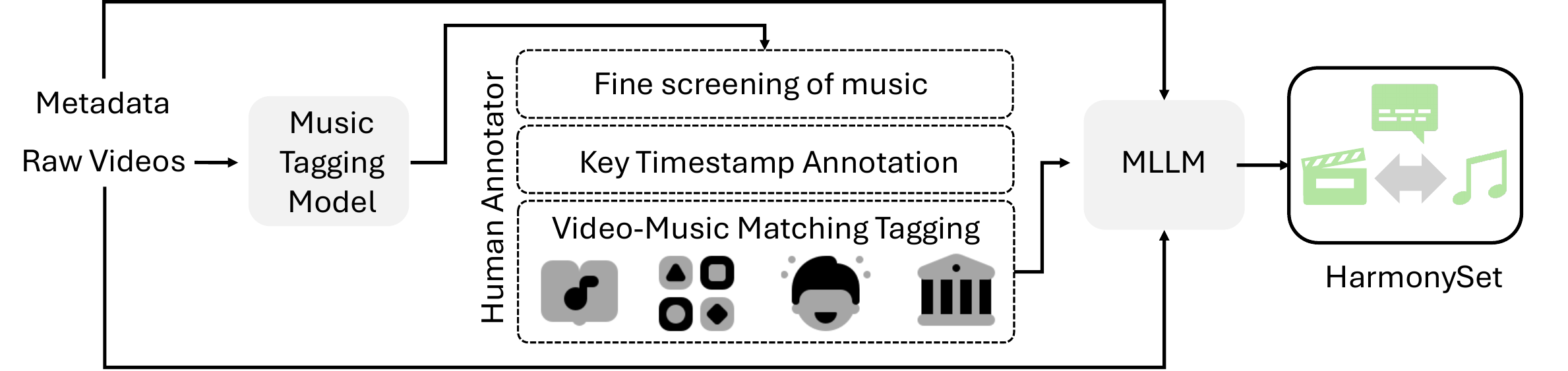} 
    \caption{Multi-phase annotation process}
    \label{fig:data_pip}
\end{figure*}

\begin{figure}[t]
    \centering
    \includegraphics[width=\linewidth]{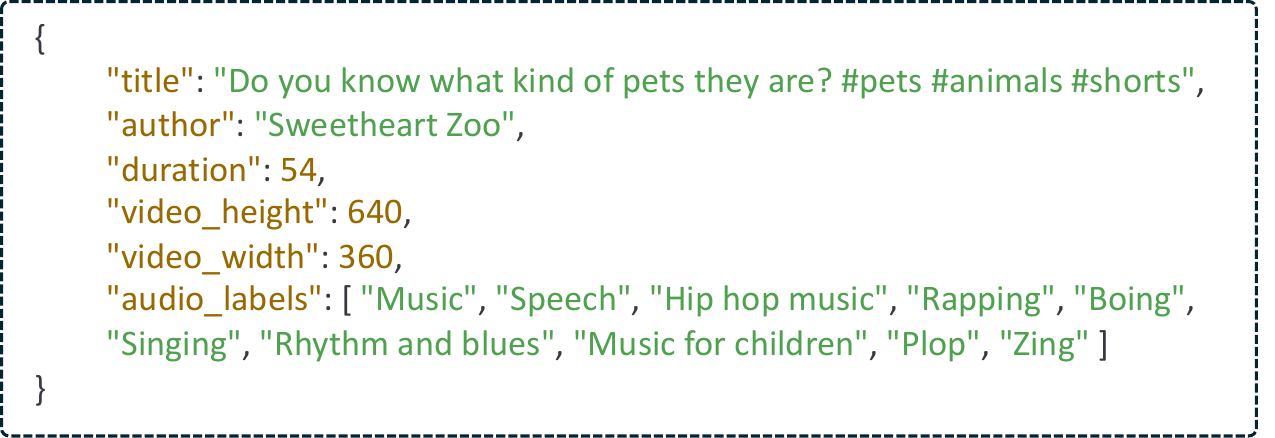} 
    \vspace{-0.5cm}
    \caption{An example of metadata of video-audio pairs. The top 10 most probable music labels generated by the PANNs are retained. Metadata used in the automated annotation generation pipeline includes video titles and music tags.}
    \label{fig:metadata}
    \vspace{-0.5cm}
\end{figure}

\begin{figure*}[t]
    \centering
    \includegraphics[width=\textwidth]{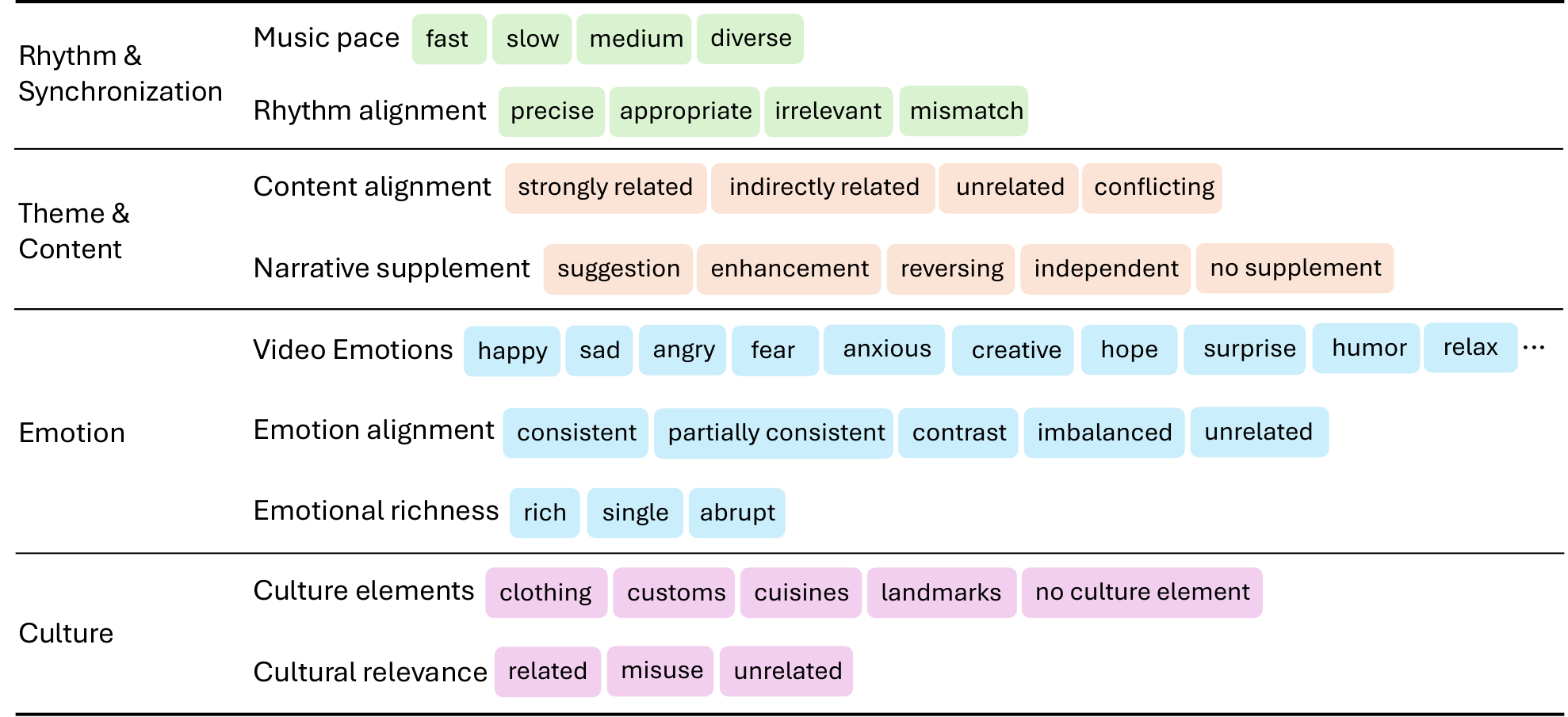} 
    \caption{A hierarchical labeling system is employed for manual annotation, encompassing four primary aspects: Rhythm and Synchronization, Theme and Content, Emotion, and Culture. These aspects are further refined into nine sub-aspects, with labels representing factual information or degrees of match.}
    \label{fig:label_list}
\end{figure*}

\begin{figure*}[t]
    \centering
    \includegraphics[width=\textwidth]{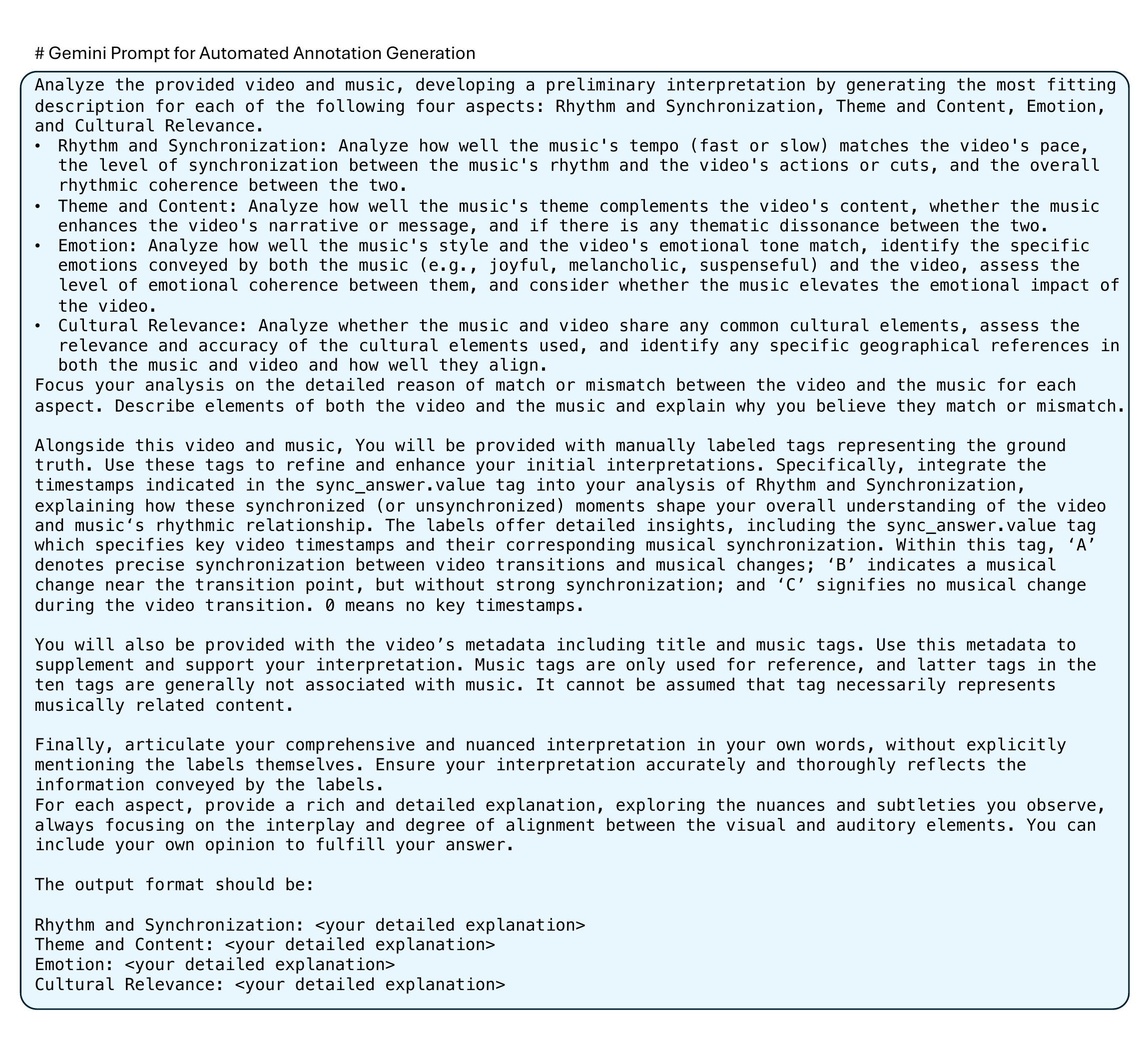} 
    \caption{Prompt for generating automated annotation using Gemini 1.5 Pro. Inputs include video and audio content, manual annotated labels, and metadata. The model is tasked with providing a detailed understanding of the match across four aspects.}
    \label{fig:gemini_prompt}
\end{figure*}

\begin{figure*}[t]
    \centering
    \includegraphics[width=\textwidth]{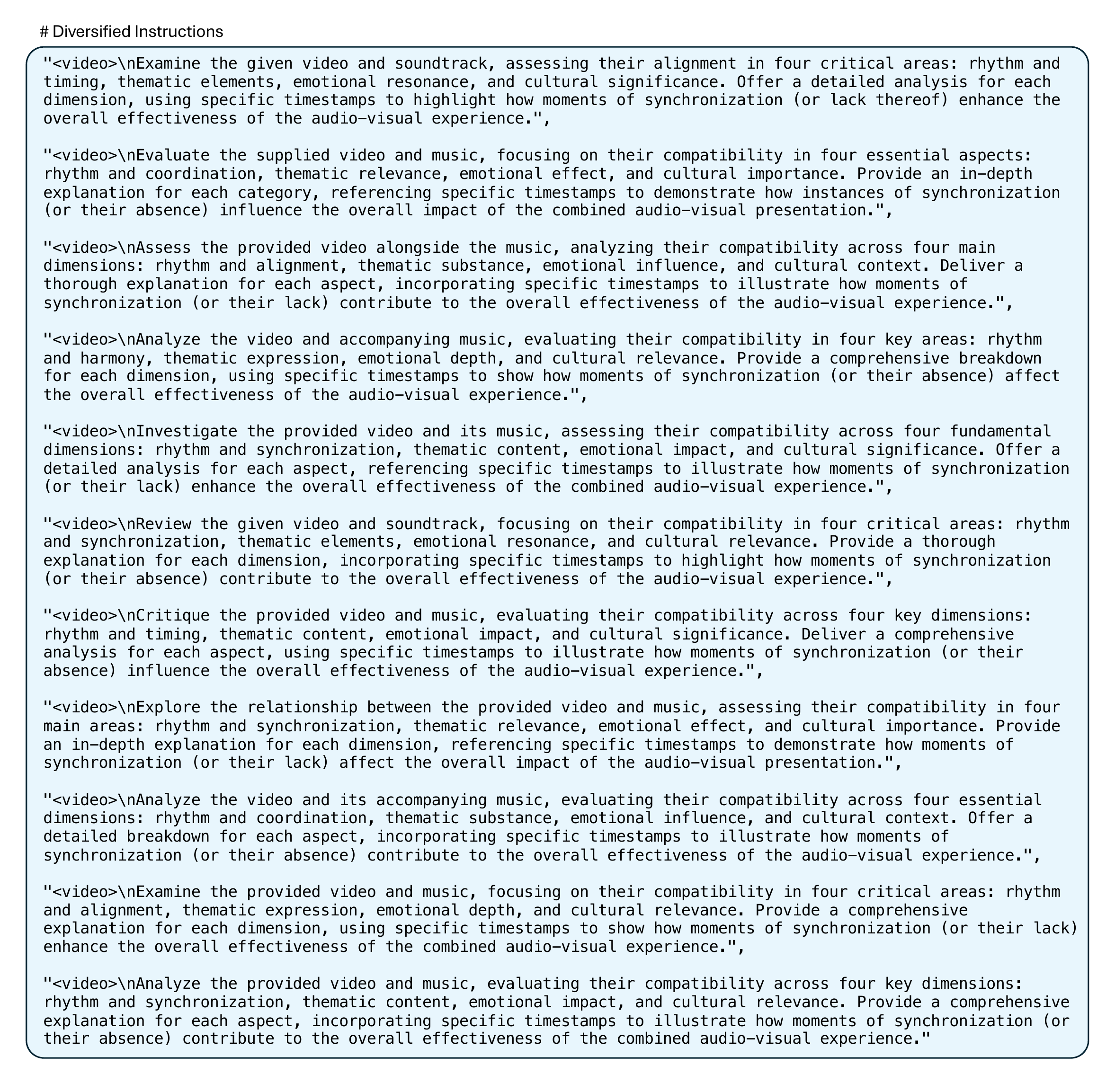} 
    \caption{Diversified instructions for HarmonySet data}
    \label{fig:instructions}
\end{figure*}

Videos in HarmonySet are sourced from the YouTube Shorts platform and were crawled using 293 keywords we designed. The complete list of keywords is shown in Figure \ref{fig:keywords}.\\
Figure \ref{fig:data_pip} illustrates the multi-phase annotation process. The raw data we crawled includes videos along with their audio and metadata. The metadata includes the title, author, duration, and video width and height, as shown in Figure \ref{fig:metadata}. We used PANNs \cite{kong2020panns} for music tagging and video filtering. The music tags generated by PANNs were added to the metadata, and videos without music were filtered out based on the tagging results. The filtering criterion was: if the labels with top 2 probabilities do not include ‘music’, the video is deleted.\\
The filtered videos were then assigned to human annotators for detailed screening to \textbf{ensure video-music pair quality} and to \textbf{exclude non-ethical and sensitive content}. The instructions for the annotators were:\\
\textit{\textbf{Music Check: }If there is no background music (e.g., pure human voice, pure environmental sound, pure noise, no sound, etc. Music mixed with human voice counts as having music), please flag the video. Listen to the entire video before making this determination, as music may only be present in a portion of the video.\\
\textbf{Content Suitability Check:} Carefully review the entire video for any content that is:
Non-Ethical: This includes, but is not limited to, content that promotes or depicts illegal activities, harmful behavior, or discrimination.
Sensitive: This includes, but is not limited to, content that is sexually suggestive, graphically violent, or exploits, abuses, or endangers children.\\
\textbf{Video Quality Check: }Please also assess the overall video quality. Flag any videos with technical issues, such as severe distortion, extremely poor resolution, or corrupted files.\\}
\textbf{Human Annotation} We conducted a rigorous annotator selection process. We recruited 120 annotators to pre-annotate 500 videos. The 120 annotators are all experts who have previous formal experience in video annotation work. After the pre-annotation, we retained 25 individuals who demonstrated both accuracy, diversity, and speed in their annotations. For videos with music, human annotators were to mark key time points and label tags. In the key time point annotation, annotators first identified moments representing visual narrative turning points or key points, then determined whether the music synchronously changed with the video at those moments. The instructions were:\\
\textit{Please mark up to three important time points in the video. If there are no changes throughout the video, fill in 0. Then determine: A) The music changes precisely in sync with the video at the turning point; B) The music changes near the turning point but is not strongly synchronized; C) The music does not change when the video turns. The answer format should be: video timestamp + comma + uppercase letter option, separated by semicolons between time points. Example: 00:10,A; 00:20,B; 00:30,C (Non-synchronization means the visual changes but the music remains the same. Examples of synchronized music changes include [outfit change on beat], [music changes to a victorious tune after a basketball shot], [music reaches a climax as the video reaches its most exciting moment], etc.)}\\
For label tagging, the structured label system is shown in Figure \ref{fig:label_list}.\\
\textbf{Automatic Annotation} After human annotation, in the automatic annotation phase, the MLLM will receive the video and audio content, human annotation results, and required metadata as input. An example of the metadata is shown in Figure \ref{fig:metadata}, where the video title and audio tags will be used in the automatic annotation process. The MLLM will generate detailed video-music alignment annotations, including semantic alignment and temporal synchronization understanding. We use Gemini 1.5 Pro as the MLLM for the automatic annotation phase, with specific instructions shown in Figure \ref{fig:gemini_prompt}. In addition, to ensure the diversity of instructions and to avoid overestimation of performance, we employed multiple prompt templates for the instruction tuning data, as illustrated in Figure \ref{fig:instructions}. These instructions convey the same underlying meaning while avoiding rigid patterns in sentence structure and word choice. The instructions in the dataset will be randomly assigned to one of these ten templates, promoting variability and enhancing the robustness of the training process. This not only helps in capturing a wider range of expressions but also mitigates the risk of the model becoming overly reliant on specific phrasing, thereby improving its generalization capabilities.\\

\subsection{More statistics}
\begin{figure*}[t]
    \centering
    \includegraphics[width=\textwidth]{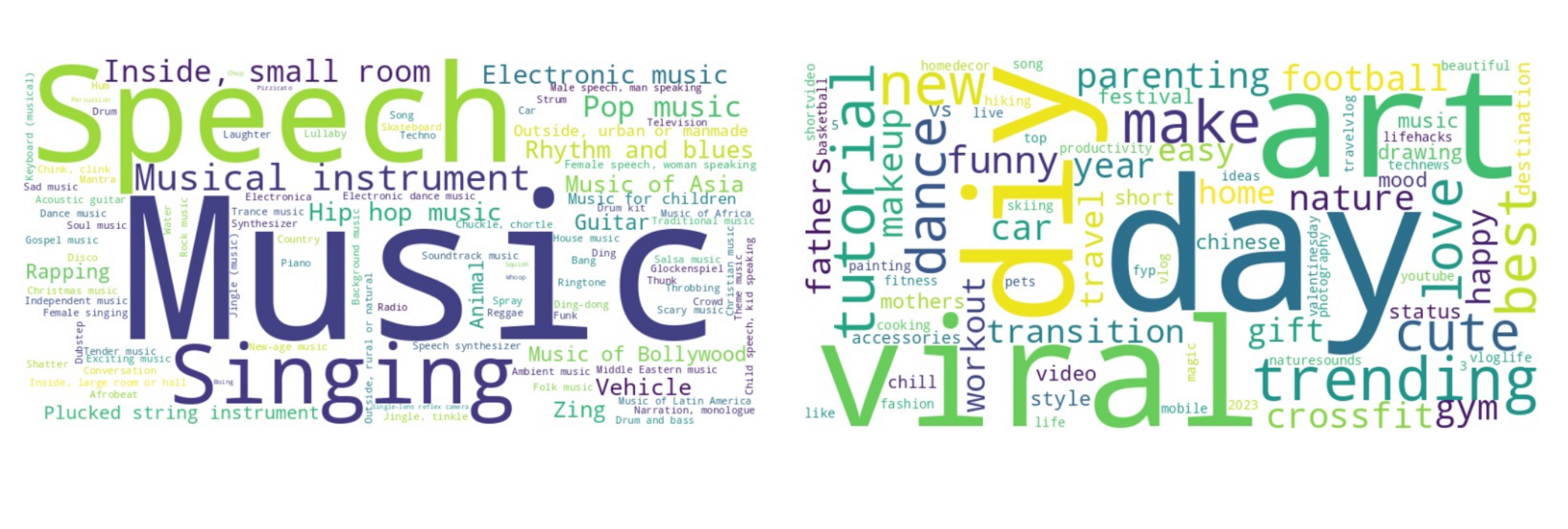} 
    \caption{(Left) Word cloud visualizations of high-frequency music tags extracted by PANNs (excluding stop words). The larger the word, the higher its frequency. This showcases the diversity of musical genres, instruments, and cultures. (Right) Word cloud visualizations of video titles (excluding stop words). The word cloud demonstrates the wide range of video types and diverse scenes included in the dataset.}
    \label{fig:wordclouds}
\end{figure*}

The raw data crawled from the platform consists of 59,771 video-music pairs. After the first round of filtering, the number of video-music pairs was reduced to 49,610. Following a meticulous manual screening, the total number of videos was further reduced to 48,328. In addition to statistics including video categories, video duration, and the number of words in the annotation text shown in Figure \ref{fig:Statistics}, we also compiled statistics on the frequency of music tags and keywords in video titles (with meaningless stop words like \textit{I} and \textit{the} removed). Figure \ref{fig:wordclouds} presents word clouds for music tags and video titles, illustrating the diversity in music genres, styles, and instruments, as well as the variety of content and themes showcased in the videos.

\subsection{Benchmark}
\textbf{Metrics of HarmonySet-OE} \\ Traditional language metrics are mainly sensitive to lexical variations and cannot identify changes in sentence semantics.
\noindent\textbf{BLEU-4}\quad BLEU \cite{papineni2002bleu}, or Bilingual Evaluation Understudy, is a metric for evaluating machine translation by comparing N-grams of the translation to human references. BLEU-4 specifically evaluates the match of four-word sequences and includes a Brevity Penalty to account for shorter translations.

\noindent\textbf{ROUGE-L}\quad ROUGE \cite{lin2004rouge}, or Recall-Oriented Understudy for Gisting Evaluation, measures the overlap between generated and reference summaries. ROUGE-L focuses on the Longest Common Subsequence (LCS), which does not require consecutive word matches.

\noindent\textbf{BLEURT}\quad BLEURT \cite{sellam2020bleurt}, or Bilingual Evaluation Understudy with Representations from Transformers, evaluates machine translation and natural language generation using pre-trained language models like BERT. It captures deep semantic information, addressing issues like synonym substitution and sentence rearrangement better than traditional metrics.

\noindent\textbf{CIDEr}\quad CIDEr \cite{vedantam2015cider}, or Consensus-based Image Description Evaluation, is designed for image captioning tasks. It uses TF-IDF weights to emphasize n-grams common in human annotations but rare in general descriptions, capturing more detailed information.

\begin{figure*}[t]
    \centering
    \includegraphics[width=\textwidth]{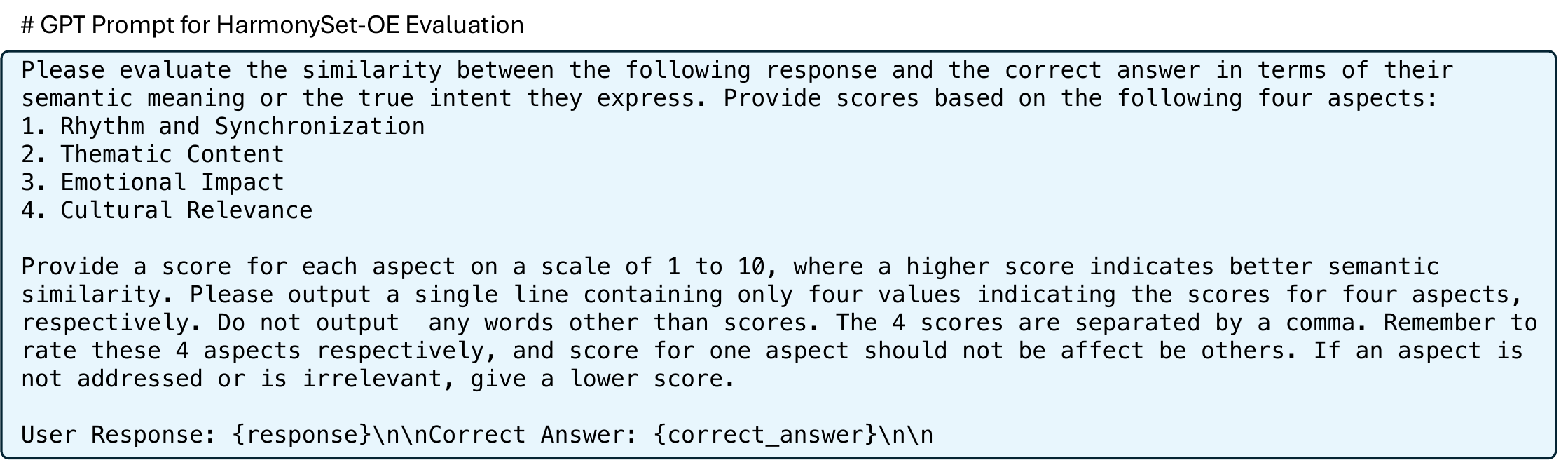} 
    \caption{Detailed prompt for using GPT4o as the evaluation metric for HarmonySet-OE. GPT4o will receive the correct answer and the model's output response and output the similarity of the response to the true answer in four aspects, assessing semantic and factual similarity rather than mere word matching.}
    \label{fig:evaluation_prompt}
\end{figure*}

\begin{figure*}[t]
    \centering
    \includegraphics[width=\textwidth]{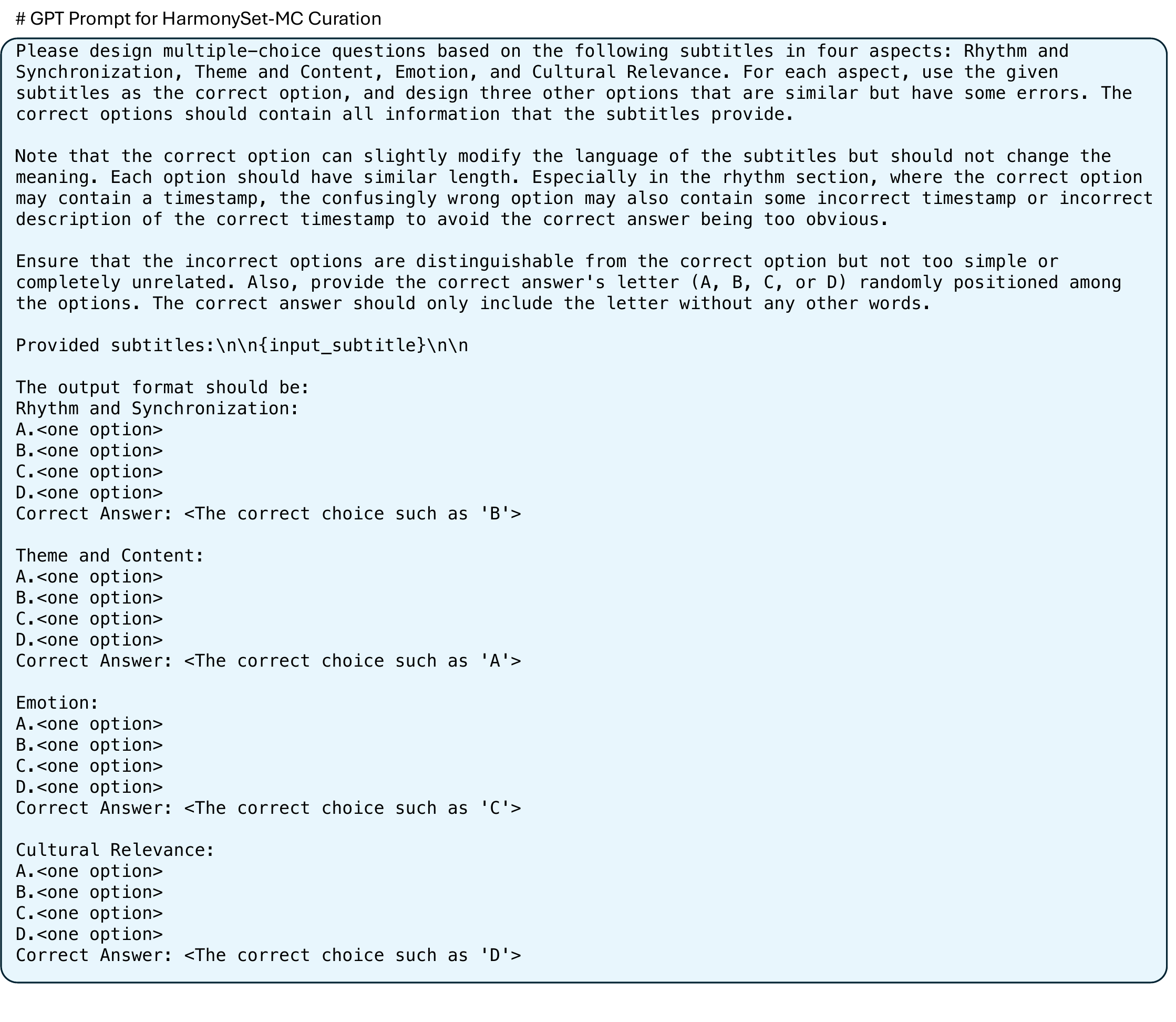} 
    \caption{Detailed prompt for generating HarmonySet-MC using GPT4o based on HarmonySet annotations. Each question includes one correct option derived from the dataset and three distractor options designed to be similar in structure, length, and theme, but containing identifiable factual errors.}
    \label{fig:mc_prompt}
\end{figure*}

\noindent\textbf{WUPS}\quad The Wu-Palmer similarity (WUPS) \cite{wu1994verb} measures the similarity of word senses based on their positions in the WordNet taxonomy. However, it struggles with words that are similar in form but different in meaning and cannot handle phrases or sentences effectively.\\
These traditional Natural Language Generation metrics lack the ability to understand and evaluate text with complex logic. In contrast, large language models are proven to be capable of comprehending text deeply. We employed GPT-4o to evaluate similarity between model responses and ground truth annotations in terms of their semantic meaning or the true intent they express. Specific prompt can be seen in Figure \ref{fig:evaluation_prompt}.\\\textbf{HarmonySet-MC Curation} We followed the QA curation pipeline of widely used multiple-choice QA benchmarks like EgoSchema \cite{mangalam2023egoschema} and VideoMME \cite{fu2024video} to design the extended benchmark HarmonySet-MC. We used a large language model to generate three incorrect answers for each annotation that is used as the correct answer. After iterating through multiple prompt versions, we ensured that the final multiple-choice questions were challenging yet reasonable, avoiding any form-based biases beyond semantics. Detailed prompt can be seen in Figure \ref{fig:mc_prompt}. For the LLM selection, we tested GPT4o, Claude, and GPT4, ultimately choosing GPT4o as the model for generating the multiple-choice questions.

\section{More details of Experiment}

\subsection{Baseline Introduction}
Our open-source model baselines include VideoLLaMA2 \cite{cheng2024videollama} and Video-SALMONN \cite{sun2024video}, both state-of-the-art video-audio multimodal large language models. Earlier MLLMs capable of processing both video and audio, such as Macaw-LLM \cite{lyu2023macaw}, are no longer available due to a lack of maintenance and therefore were not included in our experiments.\\
\textbf{Video-LLaMA2} VideoLLaMA2 is a Video Large Language
Model designed for spatial-temporal modeling and audio understanding. VideoLLaMA2 comprises a vision-language branch, an audio-language branch, a Spatial-Temporal Convolution Connector (STC Connector), and a Large Language Model (LLM). The vision-language branch uses a CLIP image encoder (ViT-L/14) to process individual frames, then aggregates these features using a novel Spatial-Temporal Convolution Connector (STC Connector) designed to preserve spatial-temporal information efficiently. The audio-language branch transforms audio into spectrograms, encodes them using BEATs, and then uses an MLP to align the audio features with the LLM. The chosen LLMs for this architecture are Mistral-Instruct and Qwen2-Instruct. The STC Connector prioritizes maintaining spatial-temporal order, minimizing token count, and mitigating information loss during downsampling through the use of 3D downsampling and RegStage convolution blocks.\\
\textbf{Video-SALMONN} Video-SALMONN is designed for obtaining fine-grained temporal information required by speech understanding. Video-SALMONN uses pre-trained encoders for visual (InstructBLIP), speech (Whisper), and non-speech audio (BEATs) inputs. These features are temporally synchronized, aligning audio and visual features at the video frame rate (2Hz). A Multi-Resolution Causal (MRC) Q-Former then processes these synchronized features at different time scales (e.g., 1, 5, and 10-second windows) to capture fine-grained audio-visual joint representations. \\

\subsection{Implementation Details}
For VideoLLaMA2 training, we utilized 4 NVIDIA H800 GPUs (a total of 32 GPUs). The training configuration followed the default settings of VideoLLaMA2-AV, except for the learning rate (lr), which was adjusted to 1e-5. The model was trained for 2 epochs on our dataset. All testing was conducted on a single NVIDIA H800 GPU (a total of 8 GPUs). Experiments in Table \ref{T:experiment} used 16 frames for testing. For experiments in Table \ref{T:sup_frame}, which explore the impact of varying frame numbers, the number of frames used for testing matched the number of frames used during training (16/32/64 frames).
\subsection{More analysis of main results}
In Table \ref{T:experiment}, we present scores for each model across six main video categories and four evaluation aspects. A horizontal comparison across the six video categories reveals that models generally score lowest on knowledge-based videos and highest on arts and performance videos. This may be because knowledge-based videos often require deeper semantic understanding and factual recall. The criteria for evaluating knowledge-based videos might be more stringent, reflecting the need for accurate information retrieval. On the other hand, the evaluation of arts and performance videos could be more subjective and open to interpretation, potentially leading to higher scores. A vertical comparison across the four evaluation aspects shows that all models consistently score lowest on the cultural aspect, suggesting that understanding and evaluating cultural nuances remains a significant challenge for current models. This could be due to the inherent complexity and subjectivity of cultural interpretations, and the current models lack sufficient training data that adequately represents the diversity and depth of cultural contexts. This deficiency hinders their ability to accurately assess culturally relevant aspects of the videos.\\
While Gemini-1.5 Pro generally performs well, the HarmonySet-enhanced VideoLLaMA2 demonstrates that open-source models can achieve comparable or even superior performance. This highlights the potential of open-source development in the MLLM domain. However, the base VideoLLaMA2 and video-SALMONN lag significantly, indicating that further research and development are needed to close the gap with closed-source models without relying on additional datasets like HarmonySet.\\
\subsection{Experiments on General Audio-Visual Tasks}

\begin{table}[t]
\caption{Performance on AVSD for General Audio-Visual QA.} 
\label{tab:avsd}
\vspace{-0.3cm}
\huge
\centering
\resizebox{\linewidth}{!}{
\begin{tabular}{lcccc}
\toprule
Metrics         & BLEU                              & BLEU-4                             & ROUGE       & BERT                         \\ \midrule
VideoLLaMA2   &  0.28  & 0.19 & 0.33 & 0.87 \\
VideoLLaMA2 (HarmonySet) & \textbf{0.30}  & \textbf{0.21} & \textbf{0.35} & \textbf{0.89}  \\
\bottomrule 
\end{tabular}
}
\vspace{-0.7cm}
\end{table}
Table \ref{tab:avsd} presents the performance of VideoLLaMA2 on AVSD dataset before and after training with HarmonySet. AVSD is a widely used dataset for audio-visual question answering. It includes general audio-visual QA tasks, such as "Do you hear any audio at all?" and "Is there a violin sound in the background of the video?" From the table, it is evident that the VideoLLaMA2 trained with HarmonySet has achieved improved performance across all metrics. This indicates that HarmonySet not only enhances the model's understanding of the relationships between video and music but also proves beneficial for conventional tasks, such as perception. We will further investigate the specific insights and capabilities that HarmonySet can provide, aiming to deepen our understanding of its impact on model performance and its potential applications in various audio-visual tasks.
\\
\subsection{More details of ablations}
\textbf{Full ablation on different training data} Table \ref{T:sup_training} shows the results of VideoLLaMA2 when not trained, trained with 10k MLLM auto-generated data, trained with 10k HarmonySet data, and trained with the entire HarmonySet data. It provides a more intuitive comparison on different types of training data, demonstrating the effectiveness and importance of incorporating human knowledge through HarmonySet.\\\begin{table}[h]
\centering
\caption{Performance of humans and models on 100 questions from HarmonySet-OE. Results show a noticeable gap between even the best model's performance and human performance, highlighting the limitations of current models in generating open-ended responses.}
\tiny
\resizebox{\linewidth}{!}{
\begin{tabular}{lcccc}
\toprule
         & R \& S                           & T                              & E    & C                       \\ \midrule
VideoLLaMA2 (HarmonySet)  &   5.49  &  5.10 &  5.25 & 4.77   \\
Human & \textbf{7.38}  & \textbf{7.02} & \textbf{7.57} & \textbf{6.32} \\
\bottomrule
\end{tabular}}
\label{T:human_performance_oe}
\end{table}\input{table/supp_training}\\\noindent\textbf{Ablation on Number of Frames} In Table \ref{T:sup_frame} we provide the results of ablation on number of frames across all six categories. Increasing frames from 16 to 32 demonstrably improves performance, highlighting the importance of sufficient temporal context. However, the performance degradation with 64 frames reveals that more frames do not necessarily translate to better results, especially for short-form videos. This suggests potential overfitting, information redundancy, or an unfavorable cost-benefit ratio regarding computational resources. Crucially, this indicates that effectively tackling our dataset's challenges doesn't require excessive computation. A moderate frame count (32 in this instance) appears to strike an optimal balance, maximizing performance while minimizing computational burden. This underscores the possibility of creating efficient and effective solutions for short-form video analysis without resorting to computationally intensive strategies, and emphasizes the importance of optimizing frame selection based on video characteristics.
\input{table/supp_frame}
\subsection{Human performance on HarmonySet-OE}
 We also evaluated both human and model performance on HarmonySet-OE, shown in Table \ref{T:human_performance_oe}. Due to the complexity of generating open-ended answers manually, we randomly selected 100 questions from HarmonySet-OE and collected answers from three different annotators per question. Human-generated answers were evaluated using the same methodology applied to the models. We compared human performance against VideoLLaMA2 (HarmonySet), the best-performing model in our main experiment. Results show a noticeable gap between even the best model's performance and human performance, highlighting the limitations of current models in generating open-ended responses. However, the performance gap between humans and models on the OE task is smaller (e.g., in the cultural aspect, the human score is 6.32, while the model score is 4.77) compared to the multiple-choice task (e.g., human accuracy on the cultural aspect is 93.81\%, while the best model accuracy is only 50.40\%). This smaller gap in open-ended responses might be attributed to the higher cost for humans to produce long-form text, whereas models can achieve higher scores by increasing text richness and length. This suggests that the OE task presents certain challenges even for humans.

%% file: table/supp_training.tex
\begin{table*}[h]
\caption{Full ablation on impact of different training data. Results reveal that training with 10,000 automatically generated annotations provides minimal performance improvement and even hinders performance on Theme and Emotion aspects, suggesting potential inaccuracies or misleading information in the auto-generated data. In contrast, training with HarmonySet data consistently enhances performance, with greater improvements observed with larger training sets. This demonstrates the effectiveness and importance of incorporating human knowledge through HarmonySet.}
\label{T:sup_training}
\vspace{-10pt}
{
\renewcommand\baselinestretch{0.9}

\normalsize
\tiny
\setlength\tabcolsep{5pt} 
\resizebox{1.0\linewidth}{!}{
\begin{tabular}{lcccccccc}
\toprule

Models    & Metrics & \begin{tabular}[c]{@{}c@{}}Life \&\\Emotion \end{tabular}  & \begin{tabular}[c]{@{}c@{}}Art \&\\Performance \end{tabular}  & \begin{tabular}[c]{@{}c@{}}Travel \&\\Events \end{tabular}  &  \begin{tabular}[c]{@{}c@{}}Sports \&\\Outdoors \end{tabular}  & Knowledge &\begin{tabular}[c]{@{}c@{}}Tech \&\\Fashion \end{tabular}  & Overall \\
\midrule

\multirow{4}{*}{VideoLLaMA2 (Vanilla)} & R \& S & 3.89 & 4.80 & 4.56 & 4.01 & 3.39 & 3.54 & 4.15          \\
 & T      & 4.09 & 4.83 & 4.93 & 3.89 & 3.44 & 3.71 & 4.29          \\
 & E      & 4.36 & 5.01 & 5.02 & 4.08 & 3.44 & 3.49 & 4.38          \\
 & C      & 2.95 & 3.46 & 3.69 & 2.56 & 2.32 & 2.52 & 3.05          \\

\midrule
\multirow{4}{*}{VideoLLaMA2 (10k, F.A.)}& R \& S & 4.56 & 5.05 & 4.97 & 4.28 & 4.20 & 4.17 & 4.59          \\
 & T      & 4.20 & 5.01 & 4.85 & 3.49 & 3.36 & 3.41 & 4.16          \\
 & E      & 4.29 & 4.76 & 4.67 & 4.03 & 3.76 & 3.79 & 4.28          \\
 & C      & 3.53 & 3.98 & 3.67 & 2.93 & 3.13 & 3.02 & 3.44          \\
 \midrule
\multirow{4}{*}{VideoLLaMA2 (10k, HamrmonySet)}& R \& S & 4.69 & 5.58 & 5.30 & 4.49 & 4.36 & 4.25 & 4.86          \\
 & T      & 4.66 & 5.02 & 4.98 & 4.40 & 4.39 & 4.29 & 4.70          \\
 & E      & 4.64 & 5.43 & 5.26 & 4.06 & 3.85 & 3.78 & 4.66          \\
 & C      & 3.99 & 4.30 & 4.25 & 2.97 & 3.79 & 3.34 & 3.89          \\
  \midrule
\multirow{4}{*}{VideoLLaMA2 (Full HamrmonySet)}& R \& S & 5.43 & 6.35 & 6.03 & 4.94 & 5.33 & 4.83 & \textbf{5.55} \\
 & T      & 5.12 & 5.21 & 5.03 & 4.84 & 5.21 & 4.85 & \textbf{5.06} \\
 & E      & 5.25 & 6.41 & 5.84 & 4.00 & 4.88 & 4.47 & \textbf{5.26} \\
 & C      & 4.87 & 4.98 & 4.72 & 3.31 & 5.23 & 4.09 & \textbf{4.62} \\
\bottomrule

\end{tabular}}
}
\end{table*}

%% file: table/supp_frame.tex
\begin{table*}[h]
\caption{Table \ref{T:frame_ablation} records the average results across six types of videos. This table presents the detailed results of the frame number ablation experiments. Experiments using VideoLLaMA2 trained with varying numbers of video frames (16, 32, and 64) show optimal performance with 32 frames. Performance degrades with 64 frames, indicating that using too many frames may lead to information redundancy and performance degradation.}
\label{T:sup_frame}
\vspace{-10pt}
{
\renewcommand\baselinestretch{0.9}

\normalsize
\tiny
\setlength\tabcolsep{5pt} 
\resizebox{1.0\linewidth}{!}{
\begin{tabular}{lcccccccc}
\toprule

Models    & Metrics & \begin{tabular}[c]{@{}c@{}}Life \&\\Emotion \end{tabular}  & \begin{tabular}[c]{@{}c@{}}Art \&\\Performance \end{tabular}  & \begin{tabular}[c]{@{}c@{}}Travel \&\\Events \end{tabular}  &  \begin{tabular}[c]{@{}c@{}}Sports \&\\Outdoors \end{tabular}  & Knowledge &\begin{tabular}[c]{@{}c@{}}Tech \&\\Fashion \end{tabular}  & Overall \\
\midrule

\multirow{4}{*}{16 Frames} & R \& S & 5.43 & 6.35 & 6.03 & 4.94 & 5.33 & 4.83 & 5.55          \\
 & T      & 5.12 & 5.21 & 5.03 & 4.84 & 5.21 & 4.85 & 5.06          \\
 & E      & 5.25 & 6.41 & 5.84 & 4.00 & 4.88 & 4.47 & 5.26          \\
 & C      & 4.87 & 4.98 & 4.72 & 3.31 & 5.23 & 4.09 & 4.62          \\

\midrule
\multirow{4}{*}{32 Frames}& R \& S & 5.47 & 6.37 & 6.07 & 4.98 & 5.39 & 4.87 & \textbf{5.59} \\
 & T      & 5.14 & 5.23 & 5.08 & 4.85 & 5.23 & 4.86 & \textbf{5.08} \\
 & E      & 5.26 & 6.44 & 5.88 & 4.05 & 4.90 & 4.50 & \textbf{5.29} \\
 & C      & 4.89 & 5.01 & 4.76 & 3.32 & 5.26 & 4.14 & \textbf{4.65} \\
 \midrule
\multirow{4}{*}{64 Frames}& R \& S & 5.38 & 6.31 & 5.97 & 4.86 & 5.27 & 4.76 & 5.49          \\
 & T      & 5.03 & 5.11 & 4.88 & 4.72 & 5.10 & 4.71 & 4.94          \\
 & E      & 5.22 & 6.36 & 5.80 & 3.95 & 4.82 & 4.40 & 5.21          \\
 & C      & 4.76 & 4.89 & 4.68 & 3.22 & 5.10 & 4.02 & 4.53          \\
\bottomrule

\end{tabular}}\par
}
\end{table*}